\documentclass[conference]{IEEEtran}
\IEEEoverridecommandlockouts
\usepackage{cite}
\usepackage{amsmath,amssymb,amsfonts,multirow,}
\usepackage{graphicx}
\usepackage[table,xcdraw]{xcolor}
\usepackage{textcomp}
\usepackage{xcolor}
\usepackage{bm}
\usepackage{ctable}
\usepackage{caption, subcaption}
\usepackage{gensymb}
\usepackage{soul}
\usepackage{algorithm}
\usepackage{algpseudocode}
\usepackage{amsmath}
\usepackage{enumitem}

\algrenewcommand\algorithmicrequire{\textbf{Input:}}

\usepackage{authblk}

\def\BibTeX{{\rm B\kern-.05em{\sc i\kern-.025em b}\kern-.08em
 T\kern-.1667em\lower.7ex\hbox{E}\kern-.125emX}}
\begin{document}

\title{D$^3$ARC: Time-Critical Distributed Disaster Detection for Asynchronous Cooperative Multi-Robot Systems}
\author{Nikolaos Koursioumpas*, Lina Magoula*, Nancy Alonistioti*, Ramin Khalili**
\\
* \emph{Dept. of Informatics and Telecommunications, National and Kapodistrian University of Athens, Greece} 
\\
** \emph{Huawei Heisenberg Research Center (Munich), Germany}
\\
\{nkoursioubas*, lina-magoula*, nancy*\}@di.uoa.gr \\
\{ramin.khalili**\}@huawei.com}
\maketitle

\begin{abstract}
Climate change is increasing the severity and unpredictability of natural disasters. In time-critical crises such as wildfires, traditional monitoring practices remain limited by coverage, cost, and personnel risk, paving the way for autonomous and adaptive monitoring solutions. Within this context, this paper introduces D$^3$ARC, an asynchronous distributed hierarchical framework for time-aware and reliable wildfire detection. D$^3$ARC integrates multiple robotic agents that cooperate under \textit{uncertainty} through distributed perception, shared situational awareness and coordinated actions.  A remote controller asynchronously decides upon each robot’s motion, while each robotic agent senses the environment and decides \textit{where} and \textit{how} to execute the wildfire detection. All robotic operations require time, and as time progresses, wildfires continue to spread, reducing the opportunity for early intervention. As such, all agents share a common objective: to detect a wildfire with a certain performance threshold as fast as possible and within a time limit. 
D$^3$ARC integrates mechanisms for safe navigation, coverage efficiency, cooperation and reliability. It introduces a \textit{forward-looking} capability that allows agents to anticipate the future by evaluating candidate strategies before execution. The framework is evaluated through realistic robotics simulations, ablation studies, and baseline comparisons, achieving an overall mission success up to 94\% with 89.4\% detection confidence. 
\end{abstract}

\begin{IEEEkeywords}
Distributed Agentic Intelligence, Cooperative Robotics, Computer Vision, Time Efficiency, Wildfire Detection
\end{IEEEkeywords}

\section{Introduction}\label{Intro}

Climate change is altering the frequency and intensity of environmental hazards, such as droughts, heatwaves, floods, and wildfires \cite{wmo2026globalclimate}. The increasing severity and unpredictability of these events pose significant risks to human life and ecosystems. Wildfires, in particular, are challenging as they can spread rapidly and uncontrollably across large, often inaccessible regions, making timely detection essential. Accordingly, the European Union Agency for the Space Programme (EUSPA) states that wildfires should be detected within 10 minutes of ignition \cite{EUSPA2024EMAidUserNeeds}.

Traditional wildfire monitoring practices primarily include ground-based inspections, fixed observation stations, and aerial patrols. These practices are often constrained by limited spatial coverage, high operational costs, and most importantly, risks to personnel operating in hazardous environments. This highlights the need for innovative technological solutions that, during such critical events, can effectively and safely monitor, perceive, act and adapt. 

In this direction, robotics is emerging as a promising technology for addressing such environmental crises \cite{Ghassemian2026}. To operate effectively, robots must act as intelligent agents, and as such be equipped with self-learning capabilities and context-aware autonomy \cite{6G-IA_2024_EuropeanVision}. However, considering the spatial scale, dynamicity and overall \textit{uncertainty} of such environments, the deployment of individual robotic agents may be insufficient. Instead, cooperative multi-robot agents can support distributed perception, shared situational awareness, and coordinated actions under rapidly evolving and time-critical conditions. These capabilities form cooperative agentic robotic intelligence as a new research direction, enabling faster response and greater operational effectiveness.

In real-world deployments, agentic capabilities are tightly coupled with the availability of computational, and communication resources. Particularly, each agent spends time for every sensing, computation, communication, and movement operation. Limited onboard computing can delay perception, and reasoning, while fluctuating network conditions introduce delays in coordination with other agents. These accumulated delays across perception, decision-making and action can directly affect a robot’s responsiveness, autonomy, and mobility range. Time awareness is therefore essential as delays may reduce the opportunity for early detection and allow a wildfire to spread further.

Considering these needs and challenges, this paper introduces D$^3$ARC, a hierarchical distributed agentic framework for cooperative robots, consisting of multiple robots and a remote controller (RC), designed for time-aware and reliable wildfire detection under \textit{uncertainty}. More specifically, the RC asynchronously controls the motion and pose of each robot, while each robotic agent decides upon \textit{where} and \textit{how} wildfire detection should be executed. A pool of AI wildfire detection models, differing in complexity and performance, is available to the robotic agents. The \textit{where} decision selects between onboard and remote execution based on resource availability and network conditions, while the \textit{how} decision selects the most suitable AI model from the available pool. When a robotic agent selects remote execution, it transmits its sensing data to the $\text{RC}$, which \textit{assists} the mission by performing wildfire detection remotely.

D$^3$ARC follows a two phase methodology that combines offline knowledge acquisition with online real-time decision-making. In the online real-time phase, D$^3$ARC integrates four mechanisms that improve safety, coverage efficiency, cooperation and mission reliability. A 360$\degree$-obstacle avoidance mechanism prevents unsafe robot motion near obstacles. A coverage efficiency mechanism rates regions based on number of visits and detection performance. This reduces redundant exploration and redirects the robots toward promising or unexplored regions. A call-for-aid mechanism promotes sensing and detection cooperation when at least one robot reports a promising region. An adaptive early stopping mechanism terminates the mission once reliable wildfire detection is sustained, avoiding unnecessary motion, time cost and overall mission duration. Finally, custom penalty functions promote feasible strategies that respect system objective and constraints. 

Another key innovation of D$^3$ARC is its \textit{forward-looking} decision-making capability. The $\text{RC}$ and robotic agents asynchronously act in a distributed manner but share a common mission reward only after their actions are completed. As such, each agent must \textit{look forward} by estimating the shared mission outcome of its candidate decisions, while the $\text{RC}$ must also account for the robots' subsequent actions. D$^3$ARC enables this through distributed neural regression models to evaluate multiple candidate strategies before action execution. The agents then greedily select the strategy expected to provide the best trade-off between detection confidence and mission duration. 
The key contributions can be summarized as follows:
\begin{itemize}
\item An asynchronous distributed hierarchical cooperative decision-making framework between multiple agents that jointly optimize mission duration and effectiveness.
\item The design and integration of mechanisms promoting navigation safety, coverage efficiency, cooperation and mission reliability in time-critical missions. 
\item A \textit{forward-looking} decision-making that evaluates possible futures by anticipating their consequences prior to action selection.
\item Realistic modeling of time for any robotic operation.
\item A real-time evaluation process using state-of-the-art robotics simulations. 
\item Integration of heterogeneous robot perception capabilities, where external factors limit the detection performance of some robots.
\item Cooperation effectiveness and robust performance, in terms of detection confidence, mission duration, and overall mission success, even under unseen scenarios and setups.
\item Comprehensive ablation studies and baseline comparisons that demonstrate the effectiveness and advantages of our proposed approach.
\end{itemize}

The rest of the paper is organized as follows. Section \ref{related_work} presents relevant state-of-the-art literature. Section \ref{system_model} provides the system model and section \ref{problem_formulation} provides the problem formulation. The proposed solution is presented in Section \ref{proposed_algorithm} and evaluated in Section \ref{performance_evaluation}. Finally, Section \ref{conclusions} concludes the paper.

\section{Related Work}\label{related_work}
This section presents an overview of the state-of-the-art. These efforts tackle different objectives related to search and rescue (SAR) missions, time-aware disaster management, and wildfire detection and tracking in robotic systems.

\textbf{SAR:}
Several works are focused on hazardous SAR scenarios. Cooperative SAR works \cite{XIONG2025100915, Aminzadeh2023} divide disaster areas among robots and optimize paths to improve coverage speed, balance workload, and reduce environmental uncertainty.  Perception-assisted SAR works \cite{Farsath10580372,XING2022102972} integrate computer vision, target positioning, and autonomous navigation to detect and localize victims from aerial imagery. Some works \cite{Horyna101007,romero2024,Medeiros9609965} propose decentralized and resource-aware approaches related to path planning, detection verification, communication management, and energy consumption under constrained or unavailable communication infrastructure.


\textbf{Time-Aware Disaster Management}: Time-aware disaster management has been studied in the context of mission planning and resource management. In mission planning \cite{BECK2018251,GHASSEMI2022103905, khanal2025,romero2024, Han8040138}, cooperative search, coverage, and task allocation are treated as optimization problems minimizing response, travel, or victim-discovery time while considering task deadlines, energy budgets, operational range, payload, fleet size, and uncertain task locations. In resource management \cite{9762674Yin,Lu8100564}, works optimize sensing duration, computation offloading, communication resources, and robot trajectories to satisfy latency constraints while reducing end-to-end delay or energy consumption.

\textbf{Wildfire Detection}: Works \cite{8331947Pham, 9504947Shrestha, Patrinopoulou2024, Julian2019} include methods for fire-front tracking and coverage through distributed control, persistent monitoring, evolutionary or reinforcement learning-based approaches. In \cite{Seraj2022}, fire propagation dynamics are considered to coordinate coverage over a planning horizon. Complementary, there are energy-aware approaches that adapt image resolution or select flight and detection parameters to reduce the energy cost of onboard wildfire detection \cite{11315192Akpomedaye,10206033Suo}.  

Table \ref{tab:sota-table} summarizes the key distinctions of D$^3$ARC compared to the most relevant works focusing on time-awareness and wildfire detection. As it can be inferred from the table, there are studies that directly or indirectly promote coverage efficiency during a mission, while some consider heterogeneous robotic perception capabilities. However, only a few works integrate neural network-based forward-looking reasoning in their decision-making. Asynchronous cooperation is limited, as the majority of the works rely on centralized, synchronized, or unspecified coordination. Most of the works do not integrate an AI-based detection model, as they assume that fire observations are already available. Beyond the characteristics explicitly reported in the table, there are no works that jointly consider mission time, detection reliability, network variability, and resource availability within a unified framework. These limitations motivate D$^3$ARC, which integrates these aspects through distributed hierarchical cooperative decision-making for fast and reliable wildfire detection under dynamic conditions. 

\begin{table}[!ht]
\centering
\resizebox{\columnwidth}{!}{%
\begin{tabular}{@{}lccccc@{}}
\toprule
\textbf{Study} & \textbf{\begin{tabular}[c]{@{}c@{}}Coverage\\ Efficiency\end{tabular}} & \textbf{\begin{tabular}[c]{@{}c@{}}Forward-Looking\\ Reasoning\end{tabular}} & \textbf{\begin{tabular}[c]{@{}c@{}}Asynchronous\\ Cooperation\end{tabular}} & \textbf{\begin{tabular}[c]{@{}c@{}}AI-based\\ Detection\end{tabular}} & \textbf{\begin{tabular}[c]{@{}c@{}}Heterogeneous\\ Perception\end{tabular}} \\ \midrule
\cite{BECK2018251, 9762674Yin} & \textbf{x} & \textbf{x} & \textbf{x} & \textbf{x} & \checkmark \\
\cite{khanal2025,Julian2019} & \checkmark & \checkmark & \textbf{x} & \textbf{x} & \textbf{x} \\
\cite{Seraj2022,8331947Pham,9504947Shrestha} & \checkmark & \textbf{x} & \textbf{x} & \textbf{x} & \textbf{x} \\
\cite{GHASSEMI2022103905} & \textbf{x} & \textbf{x} & \checkmark & \textbf{x} & \textbf{x} \\
\cite{romero2024} & \checkmark & \textbf{x} & \textbf{x} & \checkmark & \checkmark \\
\cite{Lu8100564} & \textbf{x} & \checkmark & \textbf{x} & \textbf{x} & \textbf{x} \\
\cite{Patrinopoulou2024} & \checkmark & \textbf{x} & \textbf{x} & \textbf{x} & \checkmark \\
\cite{10206033Suo} & \checkmark & \textbf{x} & \textbf{x} & \checkmark & \textbf{x} \\
\cite{11315192Akpomedaye} & \textbf{x} & \textbf{x} & \textbf{x} & \checkmark & \textbf{x} \\
\rowcolor[HTML]{EFEFEF} 
\textbf{D$^3$ARC} & \checkmark & \checkmark & \checkmark & \checkmark & \checkmark \\ \bottomrule
\end{tabular}%
}
\caption{State-of-the-art Comparison Table: Time-Awareness \& Wildfire Detection}
\label{tab:sota-table}
\end{table}
\section{System Model} \label{system_model}
This section presents the system model comprising multiple agents, a set of UAV robots $\mathcal{R}$ and a remote controller, denoted by $\text{RC}$. All agents are equipped with AI capabilities, tasked with a mission to cooperatively explore an unknown environment for potential wildfire incidents. Each robot $r \in \mathcal{R}$ is capable of sensing the environment and performing onboard wildfire detection, while the $\text{RC}$ asynchronously handles each robot’s motion and is also equipped with a wildfire detection algorithm $l_{\text{RC}}$ to assist each robot when needed.

Time is represented by asynchronous timesteps indexed by $k\in\mathbb{N}$. A new timestep occurs when one or more robots complete their assigned tasks. Let $\mathtt{t}_k \in \mathbb{R}_+$ denote the wall-clock time at which timestep $k$ starts. Any robot whose task is completed within a tolerance $\mathtt{tl}$ of $\mathtt{t}_k$ is considered \textit{ready} at the same timestep and does not trigger a new one. We denote the set of robots that are \textit{ready} at $k$ by $\mathcal{R}_k^{\mathrm{ready}}$. The duration of $k$ is the elapsed time from the beginning of $k$ to the beginning of timestep $k+1$ and is defined as $\mathtt{T}(k) = \mathtt{t}_{k+1}-\mathtt{t}_k$. The cumulative elapsed time starting from the first timestep ($k=1$) until the end of $k$ is denoted by $T_{total}(k) = \sum_{\tau=1}^{k} \mathtt{T}(\tau)=\mathtt{t}_{k+1}-\mathtt{t}_1$. When a robot $r$ becomes \textit{ready}, the RC asynchronously assigns its next trajectory and pose, denoted by $\mathcal{P}_{k,r}$, without waiting for the other robots to complete their tasks. For simplicity, trajectory and pose selection will be referred to as motion command. The execution time of this motion command varies across robots and timesteps and is denoted by $t_{k,r} \in \mathbb{R}_+$. Each robot spends time to offer a series of capabilities:

\textbf{Robot Movement Capabilities}: Each robot can move along the three axes, rotate 360$\degree$, and is able to actuate its joints to control the orientation of its onboard sensors (e.g. camera gimbal). Let $\{x_{k,r},y_{k,r},z_{k,r}\}$ denote robot's $r$ movement along axes, $\mathtt{r}_{k,r}$ denote the robot's body rotation and $\textbf{so}_{k,r}$ the orientation of all available sensors at $k$.

\textbf{Robot Sensing Capabilities}: At each timestep, each robot senses the environment for a fixed time period denoted by $\mathtt{s} \in \mathbb{R}_+$\footnote{We assume that $\mathtt{s}$ is sufficient to acquire newly generated sensing data.}. Let $\mathcal{S}_r$ denote the set of sensors available at robot $r$. Let $I_{s,r}$ denote the sensing sample rate of sensor $s \in \mathcal{S}_r$ and $w_{s,r}$ the size of each data sample. 

\textbf{Robot Computational Capabilities}: We assume time-varying computational resources at each robot. As such, let $f_{k,r}, n^{\mathtt{cpu}}_{k,r} \in \mathbb{R}_+$ denote the available computational capacity (i.e., CPU speed) and number of CPU cores of robot $r$ available at timestep $k$, and $\varsigma_r \in \mathbb{R}_+$ the CPU's effective switched capacitance (ESC). At each timestep, each robot can complete a certain number of Floating Point Operations (FLOPs) per cycle, denoted by $c_{k,r} \in \mathbb{R}_+$. 

\textbf{Robot Perception Capabilities}: Each robot perceives the environment to perform wildfire detection. A set of AI algorithms $L$ is available for each robot to select at each $k$. The computational complexity $\alpha_{l_{k,r}} \in \mathbb{R}_+$ of a selected AI algorithm $l_{k,r} \in L$, depends on the architecture and input size, and is measured in FLOPs. The average performance of an AI algorithm at $k$ is denoted by $\eta_{l_{k,r}}$ and could refer to metrics such as confidence score and accuracy. The time required by robot $r$ to process one sample using $l_{k,r}$ is denoted by $\tau_{\alpha_{l_{k,r}}} \in \mathbb{R}_+$. Regardless of the selected detection algorithm, robots may have heterogeneous detection capabilities. Their perception can be affected by external factors such as image blur, illumination variations, reflections, and viewpoint-related artifacts. Based on their ability to produce reliable detections, robots are therefore classified as either \textit{standard-detection robots} or \textit{limited-detection robots}.

\textbf{Robot Communication Capabilities}: At each $k$, each robot may request assistance from the $\text{RC}$ by offloading its sensing data. Then, the $\text{RC}$ executes $l_{\text{RC}}$ and achieves an average performance $\eta_{l_{\text{RC},k, r}}$. The communication conditions between a robot and the $\text{RC}$ are time-varying and the channel is modeled as a flat-fading with Gaussian noise power density $N_0 \in \mathbb{R}$ and channel gain $g_{k, r} \in \mathbb{R}$, where the fading is assumed constant during the timestep. Also, let $b_{k,r} \in \mathbb{R}_+$ and $p_{k,r} \in \mathbb{R}_+$ be the available bandwidth and transmission power of robot $r$, respectively. By $\mathtt{d}_{k,r} \in \mathbb{R}_+$ we denote the achievable data rate of the communication channel between the $\text{RC}$ and robot $r$ at $k$. Table \ref{table_notations} summarizes all the notations of the system model.

\begin{table}[!ht]
\centering
\resizebox{\columnwidth}{!}{%
 \begin{tabular}{|p{0.065\textwidth}|p{0.4\textwidth}|}
\hline
\cellcolor[HTML]{EFEFEF}\textbf{Parameter} & \cellcolor[HTML]{EFEFEF}\textbf{Description} \\
\hline
$k$ & The $k^{th}$ global timestep\\ \hline
$\mathtt{t}_k$ & The wall-clock time at which timestep $k$ is triggered\\ \hline
$\text{RC}$ & The AI-enabled remote controller \\ \hline
$\mathcal{R}$ & Full set of robots \\ \hline
$\mathcal{R}_k^{\text{ready}}$ & Set of robots ready to receive new $\text{RC}$ command at $k$\\ \hline
$\mathcal{P}_{k,r}$ & The motion command of robot $r$ at $k$\\ \hline
$t_{k,r}$ & The time required for robot $r$ to take an assigned pose at $k$\\ \hline
$\mathtt{s}$ & The sensing period for all robots \\ \hline 
$\beta^{\max}_r$ & Robot's $r$ maximum battery capacity \\ \hline
$\beta_{k,r}$ & Robot's $r$ available battery capacity at $k$ \\ \hline
$f_{k,r}$ & Robot's $r$ available computational capacity at $k$ \\ \hline
$n^{cpu}_{k,r}$ & The number of robot's $r$ available CPU cores at $k$ \\ \hline
$c_{k,r}$ & The number of FLOPs per cycle robot $r$ can complete at $k$ \\ \hline
$\varsigma_r$ & Robot's $r$ CPU effective switched capacitance \\ \hline
$\mathcal{S}_r$ & The set of installed robot's $r$ sensors \\ \hline
$I_{s,r}$ & The sensing sample rate of robot's $r$ sensor $s$ \\ \hline
$w_{s,r}$ & The data sample size of robot's $r$ sensor $s$ \\ \hline
$L$ & The set of AI algorithms available for all robots \\ \hline
$\alpha_{l_{k,r}}$ & The computational complexity of the AI algorithm $l_{k,r}$ at $k$\\ \hline
$l_{\text{RC}}$ & The AI algorithm at the remote controller $\text{RC}$ \\ \hline
$\alpha_{l_{\text{RC}, k, r}}$ & The computational complexity of the $\text{RC}$'s $l_{\text{RC}}$ using robot's $r$ sensing data at $k$\\ \hline
$\eta_{l_{k,r}}$ & The average performance of $l_{k,r}$ executed at $k$ \\ \hline
$\eta_{l_{RC,k,r}}$ & The average performance of $l_{\text{RC}}$ executed at $k$\\ \hline
$\tau_{\alpha_{l_{k},r}}$ & The computation time to execute $l_{k,r}$ at $k$ \\ \hline
$\mathtt{d}_{k,r}$ & Robot's $r$ achievable transmission data rate at $k$ \\ \hline
$N_0$ & The white Gaussian noise power spectral density\ \\ \hline
$g_{k,r}$ & The gain of the wireless channel robot $r$ has access to at $k$ \\ \hline
$b_{k,r}$ & Robot's $r$ allocated bandwidth at $k$ \\ \hline
$p_{k,r}$ & The transmission power of robot $r$ at $k$ \\ \hline
\end{tabular}
}
\caption{Notation Table} \label{table_notations} 
\vspace{-10pt}
\end{table}

\section{Problem Formulation}\label{problem_formulation}
Both the $\text{RC}$ and the robots are tasked with cooperatively detecting a wildfire incident with a required level of confidence. However, satisfying a detection confidence requirement alone is not sufficient, since robot operation involves time-consuming activities, including mobility, sensing, computation, and communication. The objective is to detect a wildfire incident with a certain confidence as fast as possible under an imposed time limit. 

Achieving the objective requires a chain of hierarchical, distributed, and interdependent decisions taken by both the $\text{RC}$ and the robots. Each agent controls complementary parameters and operates under a different view of the same environment, necessitating coordinated decision-making.

\textbf{Decision-Making Stage 1 (@$\text{RC}$)}: At the beginning of each $k$, the first stage of decision-making takes place at the $\text{RC}$. The $\text{RC}$ asynchronously decides upon a motion command ($\mathcal{P}_{k,r}$) for each $r \in \mathcal{R}^{\text{ready}}_k$. The assigned command will influence all subsequent robot's $r$ decisions, as motion precision affects the robot's perspective and can facilitate progress toward achieving the end goal and vice versa.

\subsection{Requirements for Robot Movement and Sensing} \label{energy_movements}

To follow the $\text{RC}$'s motion command, each robot executes multiple movements. Three motion types are considered: hovering, horizontal, and vertical movement, each with distinct time costs. More specifically: 

The total time required by each robot $r \in \mathcal{R}^{\text{ready}}_k$ to complete a motion command at $k$, denoted by $t_{k,r}$, is expressed as:

\begin{equation}
t_{k,r} = \sum_{\mathtt{m} \in \mathcal{M}} t_{\mathtt{m},k,r}, \;r\in \mathcal{R}^{\text{ready}}_k
\end{equation}

where:
\begin{equation}
  t_{\mathtt{m},k,r}  =  
\frac{\delta^{\mathtt{m}}_{k,r}}{v^{\mathtt{m}}_{k,r}},
\end{equation}

$\mathtt{m} \in \mathcal{M} = \{hovering, horizontal, vertical\}$ represents the set of movement types, $t_{\mathtt{m},k,r}$ denotes the time required by robot $r$ to complete each movement at $k$, $\delta^{\mathtt{m}}_{k,r}$ is the distance traveled by robot $r$ along movement type $\mathtt{m}$ at step $k$, and $v^{\mathtt{m}}_{k,r}$ is the corresponding speed of robot $r$. 

\textbf{Decision-Making Stage 2 (@Robot):} Following the motion command (at $t_{k,r}$) and before sensing the environment, robot $r$ decides on a binary variable $o_{k,r}$ indicating whether to offload the detection task or proceed with local execution. In the case of local execution, robot $r$ also decides on an appropriate AI algorithm $l_{k,r}$.

\begin{figure}[ht!]
 \centering
 \includegraphics[width=0.9 \linewidth]{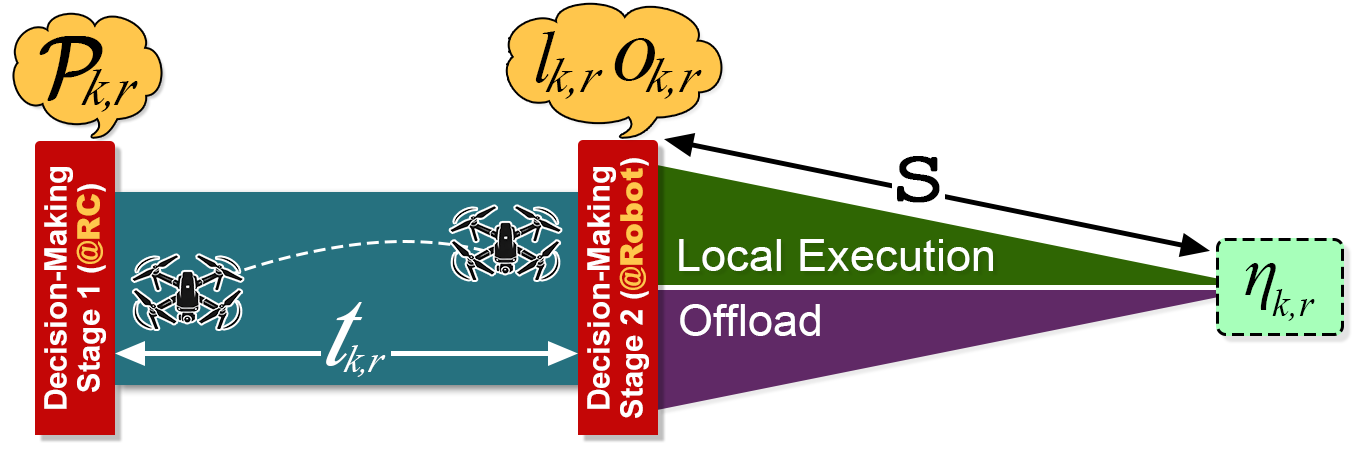}
 \caption{Robot Decision-Making Stages}
 \label{timeline_decisions}
\end{figure}

Figure \ref{timeline_decisions} illustrates the timeline of a robot $r \in \mathcal{R}^{ready}_k$, showing the chain of decisions between the RC and the robot, their dependencies, and when each decision is made. 

Each robot, after $t_{k,r}$, senses the environment for a fixed time period $\mathtt{s}$ using its set of sensors $\mathcal{S}_r$. 

\subsection{Requirements for Local Execution} \label{local_energy_detection}

In case of local execution, each robot processes the sensing data locally using the selected AI algorithm $l_{k,r}$. The computation time required by robot $r$ ($r \in \mathcal{R}^{\text{ready}}_k$) for executing $l_{k,r}$, assuming perfect parallelization across $n^{\mathtt{cpu}}_{k,r}$ cores, is denoted by $T_{computation}$ is given by:

\begin{equation}
\small
\begin{aligned}
    T_{computation}(\alpha_{l_{k,r}}) = 
     \frac{\mathtt{N}(I_{s,r}, \mathtt{s}) \cdot\alpha_{l_{k,r}}} {c_{k,r} \cdot n^{\mathtt{cpu}}_{k,r} \cdot f_{k,r}} \quad (seconds), \; 
\end{aligned}
\end{equation}
where: 
\begin{equation}
\begin{aligned}
    \mathtt{N}(I_{s,r}, \mathtt{s}) = \sum_{s \in \mathcal{S}_r} I_{s,r} \cdot \mathtt{s},
\end{aligned}
\end{equation}
denotes the number of data samples collected during the sensing period $\mathtt{s}$ to be processed by $l_{k,r}$.

\subsection{Requirements for Offloading} \label{remote_energy_detection}
In case of offloading, each robot $r \in \mathcal{R}^{\text{ready}}_k$ transmits the sensing data for remote wildfire detection. We assume that the RC has high computational capacity, and thus its processing time is negligible. Hence, only the data transmission time is considered, denoted by $T_{transmission}(k,r)$, and defined as:

\vspace{-6pt}
\begin{equation} 
\small
\label{transmission_time}
  T_{transmission}(k,r) = \frac{\sum_{s \in \mathcal{S}_r}\mathtt{N}(I_{s,r}, \mathtt{s}) \cdot w_{s,r}}{\mathtt{d}_{k,r}} \quad (seconds),
\end{equation}
where the modeling of the achievable data rate ${\mathtt{d}_{k,r}}$ at $k$ is provided in the Appendix.

\subsection{Detection Confidence} \label{ai_performance_requirements}
Each robot senses the environment over a time window $\mathtt{s}$ and performs wildfire detection. The detection confidence level may fluctuate over time due to variations in sensing quality, introducing detection uncertainty (Section \ref{system_model}, \textbf{Robot Perception Capabilities}). To capture this, we introduce a temporal consistency measure, named the Stable Confidence Score (SCS), to quantify detection stability over $\mathtt{s}$. The SCS of a robot $r$ denotes the confidence level obtained at the end of $k$ by performing either local detection ($\eta_{l_{k,r}}$), or remote detection ($\eta_{l_{\text{RC},k,r}}$), depending on the robot's decision. For simplicity, the selected score is denoted by $\eta_{k,r}$ and is defined as follows:
\vspace{-5pt}
\begin{equation}
    \eta_{k,r} = \max\Big(0, \;\overline{cs}_{{k,r}} \cdot (1 - \frac{\sigma_{{cs_{{k,r}}}}}{\overline{cs}_{{k,r}}})\Big), 
\end{equation}
where $\overline{cs}_{{k,r}}$ denotes the average confidence over $\mathtt{s}$, and $\sigma_{cs_{{k,r}}}$ represents the corresponding standard deviation. The ratio $\frac{\sigma_{cs_{{k,r}}}}{\overline{cs}_{{k,r}}}$ is the coefficient of variation, measuring dispersion around the mean. 

\begin{figure}[ht!]
 \centering
 \includegraphics[width=0.9 \linewidth]{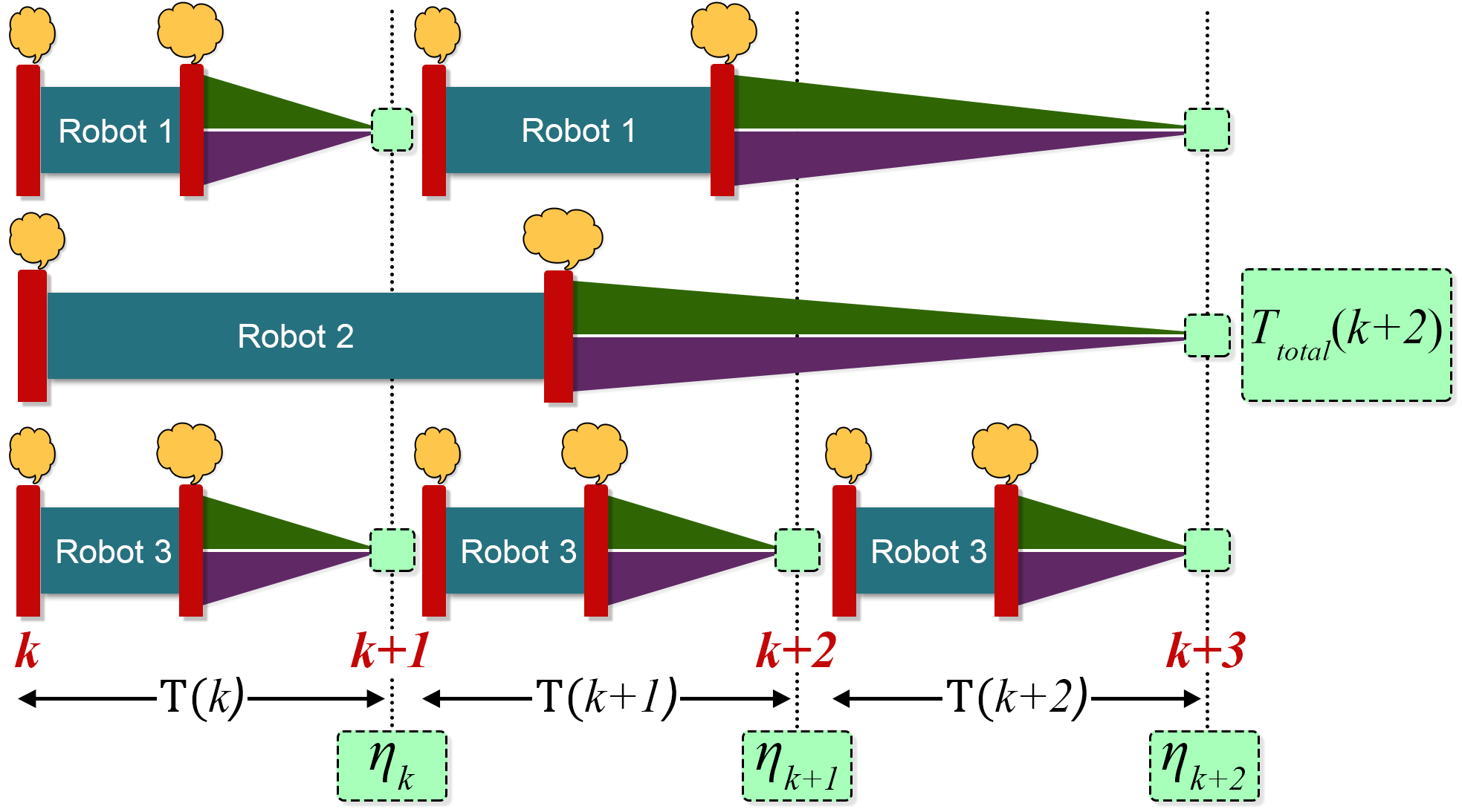}
 \caption{Illustrative Timeline Excerpt for 3 Robots}
 \label{timeline_decisions_MR}
 \vspace{-5pt}
\end{figure}

As such, at the end of $k$, individual detection confidences are combined to compute the cooperative detection confidence $\eta_k$ as follows:
\begin{equation}
\label{cooperative_conf}
\eta_k =  \max_{i \in \mathcal{R}^{\text{ready}}_k \cup \{RC\}}\eta_{k,i}.
\end{equation}
Figure \ref{timeline_decisions_MR} shows an example of 3 robots across an timeline excerpt of 4 asynchronous timesteps. The cooperative confidence $\eta_k$ is computed from Robots 1 and 3 at $k$, Robot 3 at $k+1$, and all robots at $k+2$. The total duration until the end of $k+2$ is $T_{total}(k+2) = \sum_{\tau=1}^{k+2}\mathtt{T}(\tau)=\mathtt{t}_{k+3}-\mathtt{t}_1$ .

\subsection{Problem Objective}\label{time_aware_objective}
The objective is to detect a wildfire with a certain confidence as early as possible under a time limit requirement. As such, the optimization aims to minimize the total mission time $T_{total}(k^*)$, where $k^*$ is the first timestep at which the cooperative detection confidence $\eta_{k^*}$ reaches or exceeds the required confidence threshold $\eta_{\min}$. The resulting formulation is given below:
\vspace{-5pt}
\begin{equation} \label{overall_objective}
\min_{\{\mathbf{\text{P}}_k,\mathbf{o}_k,\mathbf{l}_k\}_{k=1}^{k^*}}
T_{total}(k^*)
\end{equation}
where
\begin{equation}
    k^* = \min\{k\;|\; \eta_k \ge \eta_{\min} \}
\end{equation}
\begin{equation}
    T_{total}(k^*) = \sum_{\tau=1}^{k^*} \mathtt{T}(\tau) = \mathtt{t}_{k^*+1} - \mathtt{t}_1,
\end{equation}
\begin{eqnarray} 
    \textrm{s.t.}& \eta_{k^*} \ge\eta_{\min}, \label{performance_bound}
\end{eqnarray}
\begin{equation}
\mathtt{p}^{\min}_{k,i,r} \leq \mathtt{p}_{k,i,r} \le \mathtt{p}^{\max}_{k,i,r}, \;\forall \mathtt{p}_{k,i, r} \in \mathcal{P}_{k,r},\; \forall i \in \{1,\ldots,|\mathcal{P}_{k,r}|\}, 
\label{hardware_bound}
\end{equation}
\vspace{-17pt}
\begin{eqnarray} &&
T_{total}(k^*) \leq t_{thr}.\label{time_bound}
\end{eqnarray}

The vectors $\mathbf{\text{P}}_k = (\mathcal{P}_{k,r})_{r \in \mathcal{R}^{\text{ready}}_k}$, $\mathbf{o}_k = (o_{k,r})_{r \in \mathcal{R}^{\text{ready}}_k}$ and $\mathbf{l}_k = (l_{k,r})_{r \in \mathcal{R}^{\text{ready}}_k}$ denote the motion commands selected by the RC, and the decisions made by the robots during $k$. Constraint (\ref{performance_bound}) ensures a minimum acceptable cooperative performance. Constraint (\ref{hardware_bound}) ensures that the assigned motion command is bounded by the hardware specifications of robot $r$. Constraint (\ref{time_bound}) ensures that the elapsed wall-clock mission time up to timestep $k^*$ does not exceed the imposed threshold $t_{thr}$.

To enforce the constraints and accelerate execution, we introduce a penalty function for violations of the minimum detection confidence, motion bounds, and duration of each timestep. The penalty is defined as follows:
\vspace{-4pt}
\begin{equation}
\begin{array}{@{}l@{}}
P_{constr}(k)
= \max\bigl(0,\;\eta_{\min}-\eta_k\bigr) \\[2pt]
+\displaystyle\sum_{r\in\mathcal{R}^{\text{ready}}_k}
 \sum_{i=1}^{|\mathcal{P}_{k,r}|}
 \max\bigl(0,\;p_{k,i,r}-p^{\max}_{k,i,r},
 p^{\min}_{k,i,r}-p_{k,i,r}\bigr),
\end{array}
\label{eq:constraint_penalty_TA}
\end{equation}
\vspace{-4pt}
\begin{equation}
\begin{aligned}
P_T(k) = \mathtt{T}(k)
\end{aligned}
\label{eq:penalty_time}
\end{equation}

Overall, the complete objective function that takes into account any introduced penalties is defined as:
\begin{equation}
\label{complete_obj}
\small
\min_{\{\mathbf{\text{P}}_k,\mathbf{o}_k,\mathbf{l}_k\}_{k=1}^{k^*}}
\big(\mu_1 \cdot T_{total}(k^*) + \sum_{k=1}^{k^*}(\mu_2 \cdot P_{constr}(k) + \mu_3 \cdot P_T(k))\big)
\end{equation}
where $\mu_1, \; \mu_2 \; \text{and} \; \mu_3$ are the weights of the equation, controlling the impact of each term. 
It should be noted that each term of the objective and penalty functions is min-max normalized and then divided by its mean to ensure comparable average contributions before applying the objective weights.

\section{D$^3$ARC: Distributed Hierarchical Intelligence for Cooperative Robots}\label{proposed_algorithm}
This section introduces D$^3$ARC, an asynchronous distributed hierarchical cooperative decision-making framework designed to solve our time-aware optimization problem. The framework comprises two phases: offline warmup and online real-time execution. The $\text{RC}$ and the robots have different views of the environment and take different yet interdependent actions that are jointly evaluated using  a reward function. 

\subsection{State, Action, Reward}
\textbf{State of the Environment:} Each agent has a different, yet complementary, view of the environment, defined as the state ($S$). The $\text{RC}$ maintains a higher-level perspective as it does not directly interact with the environment, whereas the robots operate with a concrete, interaction-driven view.

\underline{$\text{RC}$ State}: The $\text{RC}$, at each timestep, inspects a set of variables $S_{\text{RC},k}$ that guides its decision-making. This set includes the robot’s position along the three axes ($x_{k-1,r}, y_{k-1,r}, z_{k-1,r}$) and pose, i.e. body's and sensors' orientation ($\mathtt{r}_{k-1,r} \; \text{and} \; \textbf{so}_{k-1,r}$), its distance from robot $r$ ($\delta_{k-1,\text{RC}, r}$), as well as the cooperative detection confidence SCS ($\eta_{{k-1}}$) and elapsed time ($T({k-1})$) observed at the previous timestep, as an outcome of the preceding joint decisions of the involved agents. Overall:
    \[
    S_{\text{RC},k} =
    \begin{aligned}
    \{\, &x_{k-1,r}, y_{k-1,r}, z_{k-1,r}, \mathtt{r}_{k-1,r}, \textbf{so}_{k-1,r},\\
    &\delta_{k-1, \text{RC},r}, \eta_{{k-1}}, \mathtt{T}({k-1}) \,\}.
    \end{aligned}
    \]
        
\underline{Robot State}: Each robot after executing RC's motion command, inspects its own set of parameters $S_{k,r}$. These parameters include its pose ($\mathtt{r}_{k,r}, \textbf{so}_{k,r}$), the cpu frequency ratio ($f^{\text{ratio}}_{k,r}$), its distance from the $\text{RC}$ ($\delta_{k,\text{RC},r}$), the allocated bandwidth ($b_{k,r}$), the achievable transmission data rate ($\mathtt{d}_{k,r}$), as well as the SCS ($\eta_{{k-1}}$) and elapsed time ($\mathtt{T}({k-1})$) observed at the previous timestep, as an outcome of the preceding joint decisions of the involved agents. Overall:
\vspace{-5pt}
\begin{align}
    S_{k,r} =
    \{\, &\mathtt{r}_{k,r}, \textbf{so}_{k,r}, f^{\text{ratio}}_{k,r}, \\&\delta_{k,\text{RC},r}, b_{k,r}, \mathtt{d}_{k,r}, \eta_{{k-1}}, \mathtt{T}(k-1)\,\}, \notag
\end{align}
    where:
    \begin{equation}
        f^{\text{ratio}}_{k,r} = \frac{f_{k,r}}{f^{\max}_r},
    \end{equation}
    and $f^{\max}_r$ denotes the maximum CPU frequency of robot $r$.

\textbf{Action Space}: The action space, is comprised of all decisions made by all agents at timestep $k$. 

\underline{$\text{RC}$ Actions}: As $A_{\text{RC},k,r}$ we denote the action space of the $\text{RC}$, that represents assigned actions related to the movement of robot $r$ along the three axes (${\delta^{(x)}_{k,r}, \delta^{(y)}_{k,r}, \delta^{(z)}_{k,r}}$), its body's and sensors' rotation ($\delta^{(\mathtt{r})}_{k,r}, \bm{\delta}^{\textbf{(so)}}_{k,r}$). Overall:
\begin{align}
A_{\text{RC},k,r} = \{\delta^{(x)}_{k,r}, \delta^{(y)}_{k,r}, \delta^{(z)}_{k,r}, \delta^{(\mathtt{r})}_{k,r}, \bm{\delta}^{\textbf{(so)}}_{k,r}\}.\notag
\end{align}

\underline{Robot Actions}: As $A_{k,r}$ we denote the decision-making of each robot $r$ regarding the offloading of the detection task $o_{k,r}$ and the model selection $l_{k,r}$. Overall:
\begin{align}
    A_{k,r} = \{o_{k,r}, l_{k,r}\}. \notag
\end{align}

\textbf{Mutual Reward}: The reward retrieved at the end of each timestep is mutual for all agents and is derived from the corresponding optimization objective (Eq. (\ref{complete_obj})). The reward is designed to promote the main objective while discouraging infeasible or undesirable decisions. As such, the mutual reward is defined as:
\vspace{-5pt}
\begin{equation}\label{reward_confidence}
rr_k = \mu_1 \cdot \eta_k -\mu_2 \cdot P_{constr} (k) -\mu_3 \cdot P_T(k).
\end{equation}

Thus, maximizing $rr_k$ directly promotes higher cooperative detection confidence towards reaching the imposed threshold as fast as possible, while penalizing constraint violations associated with the selected actions.

\subsection{Offline Warmup Phase}
This phase enables all agents to acquire initial experience, and prevents a cold-start. During warmup, and at each timestep, the RC and ready robots randomly select actions, evaluate the resulting decisions using $rr_k$, and store the corresponding states, actions, and rewards until sufficient training data are collected.

\textbf{Agent Forward-Looking and Training}: Observing the timeline within a timestep $k$ (Figure \ref{timeline_decisions}), a pair of agents have to take sequential and complementary actions at different times. However, a single shared reward is retrieved only at the end of $k$, forcing each agent to \textit{look forward} by estimating the shared outcome of its actions before the complete decision sequence is executed. Most importantly, the $\text{RC}$, lacking knowledge of each robot’s subsequent actions, evaluates the outcome of its own decision while anticipating the expected effect of the robot’s subsequent decisions on the final shared reward. This \textit{forward-looking} is the ability of all agents to anticipate the consequences of alternative actions by estimating outcomes that have not yet been observed at decision time \cite{Luo2024}.

To enable \textit{forward-looking}, one regression feedforward neural network (RFNN) model is trained for the RC and one for all robots in a supervised manner. Each model learns to estimate the resulting shared reward (model output) using as input features the agent’s respective state and actions, i.e. $S_{k,r} \cup A_{k,r}$ for all robots and $S_{\text{RC},k,r} \cup A_{\text{RC},k,r}$ for the $\text{RC}$. The rewards collected during this phase serve as ground truth for training. These trained models are used during the online phase for real-time execution.

\subsection{Online Real-Time Execution Phase}

This phase leverages the experience obtained during the offline training to enable cooperative, real-time, chained decision-making between agents. The agents use their pre-trained RFNN models and follow a greedy-based strategy. At each $k$, the $\text{RC}$ evaluates a set of $N$ candidate alternative future actions, which are sampled independently from a uniform distribution over the feasible action space\footnote{In large action spaces, exhaustive evaluation is infeasible, and as such a limited candidate set enables efficient local search.}, while each robot evaluates all possible actions. As such, each agent exploits its acquired \textit{forward-looking} knowledge: it estimates the anticipated reward for each candidate action, and greedily selects the one that maximizes this reward.

To complement and accelerate cooperative decision-making during the online phase, D$^3$ARC incorporates four additional mechanisms that support safe operation while promoting exploration, cooperation and system efficiency. These mechanisms are detailed below.

\textbf{Obstacle Avoidance Mechanism (@Robot):} The monitored environment is not obstacle-free. While the RC maintains a higher-level view of the environment, each robot has access to a more concrete local perspective and can detect nearby obstacles through onboard sensing, e.g., LiDAR. As such, it would be impractical for a robot to blindly follow every RC's motion command. D$^3$ARC incorporates a 360$\degree$ obstacle avoidance mechanism at the robot side to ensure safe and reliable exploration of the surrounding environment. This mechanism suppresses motion commands that would drive the robot toward obstacles located within a minimum safety distance. Specifically, when an obstacle is detected along the direction of motion and its distance from the robot falls below the minimum safety threshold $\mathtt{sd}_{\text{min}}$, the robot autonomously halts movement in this direction.

\textbf{Coverage Efficiency Mechanism (@RC):} Since, the RC greedily selects among a number of candidate actions at each $k$, it may repeatedly select neighboring motion commands confining the robot's motion to a nearby vicinity. To enhance environment coverage efficiency, D$^3$ARC introduces a coverage efficiency mechanism at the RC side, driven by $\eta_{k}$. The environment is partitioned into fixed-size cubic regions of size \textit{crs}. As each robot $r$ navigates across these regions, the RC maintains a region-wise logging history  $h_{cr, k}$, which records the number of robot visits $\text{vs}_{cr,k}$, and maximum SCS $\eta^{max}_{cr,k}$ observed across visits until $k$. Based on this history, each region is categorized as: 
\begin{itemize}
    \item \textbf{\textit{Unexplored}}: Unvisited regions ($cr \notin h_{cr,k}$)
    \item \textbf{\textit{Uninformative}}: Visited regions with zero average SCS ($cr \in h_{cr,k} \text{ , }\eta^{\max}_{cr,k}=0$)
    \item \textbf{\textit{Promising}}: Visited regions with non-zero maximum SCS ($cr \in h_{cr,k} \text{ , } \; \eta^{\max}_{cr,k} > 0$)
\end{itemize} 
To improve coverage efficiency, the RC assigns an area prioritization penalty $AP_{cr,r}(k)$ to each candidate action for robot $r$. This penalty is based on the history of the region that will be reached following that action. The penalty is defined as:
\begin{equation}
\small
\begin{array}{@{}l@{}}
AP_{cr,r}(k)=
\left\{
\begin{array}{@{}ll@{}}
1,
&
cr \notin h_{cr,k},
\\[8pt]

\text{vs}_{cr,k}\cdot\rho,
&
\begin{aligned}
cr &\in h_{cr,k},\\[-1pt]
\eta^{\max}_{cr,k} &= 0,
\end{aligned}
\\[12pt]

\max\left(0,\eta_{\min}-\eta^{\max}_{cr,k}\right),
&
\begin{aligned}
cr &\in h_{cr,k},\\[-1pt]
\eta^{\max}_{cr,k} &> 0.
\end{aligned}
\end{array}
\right.
\end{array}
\end{equation}
where $\rho \ge1$ controls how strongly repeated visits to \textit{uninformative} regions are penalized. As such the reward function is updated as follows:
\begin{equation}\label{reward_time_complete}
\begin{aligned}
&rr_{k} = \mu_1 \cdot \eta_k -\mu_2 \cdot P_{constr}(k)\\ & - \mu_3 \cdot P_T(k)- \mu_4 \cdot \sum_{r \in \mathcal{R}^{\text{ready}}_k}AP_{cr,r}(k),
\end{aligned}
\end{equation}
where $\mu_4$ is a weight controlling the impact of the area prioritization term. 
Since RC aims to minimize the $AP_{cr,r}(k)$ during candidate action selection, \textit{promising} regions with high previously observed SCS receive the lowest penalties and are prioritized first. \textit{Unexplored} regions receive an intermediate penalty, while repeatedly visited \textit{uninformative} regions receive increasingly larger penalties. Once again, each term is normalized according to its own scale.

\textbf{Call-for-Aid (@Robot):} To further improve cooperative sensing around \textit{promising} regions, D$^3$ARC incorporates at the robot side a Call-for-Aid (CFA) mechanism. This mechanism directs nearby robots toward a common region of interest, allowing them to collect complementary observations and jointly reduce detection uncertainty. More specifically, a robot broadcasts a CFA message to the RC and all neighboring robots within a communication radius ($\delta_{CFA}$), when it detects a wildfire with the highest confidence observed so far, yet insufficient to declare a detection. This confidence must exceed a threshold $\eta_{cfa}$ ($\eta_{cfa} < \eta_{\min}$). This threshold ensures that the CFA is triggered only when the detection confidence is high enough, preventing false interruptions to other robots’ exploration and avoiding unnecessary time loss. The broadcast message includes the position and body rotation associated with the confidence observation.

After receiving the CFA message, the RC stops providing motion commands to the CFA robots to avoid conflicting with their cooperative actions, while relaying information about the \textit{promising} region to robots that may not have received the original broadcast. CFA robots determine their subsequent movement based on the reported high-confidence position and body rotation. Robots located too far from the region, for which participation would introduce excessive travel time, remain under the RC's control and continue their current operations. Each CFA robot evaluates a set of $N$ candidate actions, sampled once again independently from a uniform distribution over the feasible action space. The evaluation favors actions that move the robot toward the \textit{promising} region, reduce its distance to the reported position, and align each robot's body rotation with that rotation of the robot that broadcast the CFA message.

To promote position and rotation alignment with the robot reported the CFA message $r_{\text{ref}}$, a new penalty $P_{CFA}(k,r,r_{\text{ref}})$ is introduced during CFA, defined as follows:
\begin{equation}
   P_{CFA}(k,r,r_{\text{ref}}) = \lambda_1 \cdot \delta_{k,r, r_{\text{ref}}} + \lambda_2 \cdot al_{k,r,r_{\text{ref}}} + \lambda_3 \cdot \mathtt{y}_{k,r,r_{\text{ref}}},
\end{equation}
where $\delta_{k,r, r_{\text{ref}}}$ is the distance between the robot's $r$ candidate position and the reported position by robot $r_{\text{ref}}$, $al_{k,r,r_{\text{ref}}}$ is the alignment between the candidate motion direction and the direction toward the reported position, and $\mathtt{y}_{k,r,r_{\text{ref}}}$ is the rotation distance between the robot's candidate rotation assignment and the reported rotation by robot $r_{\text{ref}}$. The weights $\lambda_1$, $\lambda_2$ and $\lambda_3$ are fixed design parameters that balance the relative contribution of distance, direction, and rotation alignment, respectively. Overall, lower values of $P_{CFA}(k,r,r_{\text{ref}})$ correspond to actions that better support the CFA request. 

During CFA, participating robots temporarily prioritize reaching the \textit{promising} region over sensing at intermediate locations. Accordingly, the robots increase their motion speed and reduce their sensing duration by a factor $\mathtt{fv}_{CFA}$. This allows them to reach the \textit{promising} region faster while still sensing the environment along the way, in case a better viewpoint of the event is found. Once a robot reaches the \textit{promising} region, it exits the CFA mechanism individually and returns to normal RC controlled operation, restoring its normal motion speed and sensing duration. The CFA mechanism terminates when all participating robots have reached the region.

\textbf{Adaptive Early Mission Completion Mechanism (@RC)}: Even upon wildfire detection, the RC may select actions, which can cause abrupt perspective changes and unnecessary motion. To counteract this, D$^3$ARC employs an adaptive early mission completion mechanism, improving efficiency through motion stabilization or mission termination, when the cooperative SCS indicates that further motion is unnecessary. Specifically, under reliable detections by at least one robot (line \ref{line:reliability_check}), the RC commands the respective robots to hover in place while continuing their perception tasks. Other robots, remain under normal RC control and continue their assigned tasks. If at least one hovering robot maintains a confidence greater than or equal to $\eta_{\min}$ for $esp$ consecutive timesteps, the wildfire detection is considered confirmed and the mission is successfully terminated for all robots (line \ref{line:termination_criterion}). For intermediate confidence levels (line \ref{line:intermediate_reliability}), the RC reduces the action value ranges $\mathcal{A}_{\text{RC},k,r}$ by a fixed value $\text{fv}_{a} \in \textbf{fv}, \forall a \in \mathcal{A}_{\text{RC},k,r}$ (line \ref{line:decrease_factor}). This limits abrupt changes in robot $r$'s perspective and encourages finer exploration around the current viewpoint, which may contain useful evidence of a wildfire. If intermediate confidence persists for $c_{\mathrm{thr}}$ consecutive timesteps, this indicates that the finer exploration has not produced stronger evidence. Therefore, more drastic corrective measures are applied by expanding the action set using the same value (line \ref{line:increase_factor}), allowing the robot to explore more diverse viewpoints and potentially reduce this uncertainty. If the cooperative SCS falls below the intermediate confidence threshold, the detection is no longer considered sufficient. As such, the action value ranges are reset and standard D$^3$ARC operation resumes (line \ref{line:dcra_resume_actions}).

\begin{algorithm}[!ht]
\small
\captionsetup{labelformat=empty}
\caption{\textbf{Algorithm}: D$^3$ARC Adaptive Early Mission Completion}
\label{alg:early_stopping}
\begin{algorithmic}[1]
\Require $\eta_{\min}, \eta_{\mathrm{int}}, esp, c_{\mathrm{thr}}, \textbf{fv}$
\State Initialize $c_{\mathrm{stop},r}\gets0$, $c_{\mathrm{low,r}}\gets0$
\For{each timestep $k$}
    \If{$\eta_{k-1,r}\geq\eta_{\min}$} \label{line:reliability_check}
        \State \textit{hover}; $c_{\mathrm{stop,r}}\gets c_{\mathrm{stop,r}}+1$; $c_{\mathrm{low,r}}\gets0$ 
        \If{$c_{\mathrm{stop,r}}\geq esp$} \label{line:termination_criterion}
            \State \textit{complete mission}
        \EndIf

    \ElsIf{$\eta_{\mathrm{int}}\leq\eta_{k-1,r}<\eta_{\min}$} \label{line:intermediate_reliability}
        \State $c_{\mathrm{stop,r}}\gets0$; $c_{\mathrm{low.r}}\gets c_{\mathrm{low,r}}+1$

        \If{$c_{\mathrm{low,r}} < c_{\mathrm{thr}}$}
             \State $\mathcal{A}_{\text{RC},k,r}\gets\{a/\mathrm{fv}_a\mid \mathrm{fv}_a \in \textbf{fv}, a\in \mathcal{A}_{\text{RC},k,r}\}$
             \label{line:decrease_factor}
        \Else
            \State $\mathcal{A}_{\text{RC},k,r}\gets\{a + \mathrm{fv}_a\mid \mathrm{fv}_a \in \textbf{fv}, a\in \mathcal{A}_{\text{RC},k,r}\}$ \label{line:increase_factor}
        \EndIf

    \Else
        \State \textit{reset} $\mathcal{A}_{\text{RC},k,r}$; $c_{\mathrm{stop,r}}\gets0$; $c_{\mathrm{low,r}}\gets0$ \label{line:dcra_resume_actions}
    \EndIf
\EndFor
\end{algorithmic}
\end{algorithm}
\section{Performance Evaluation} \label{performance_evaluation}
\textbf{Mission}: The mission incorporates up to 4 UAV robots. Each UAV robot is equipped with a set of sensors for wildfire detection and safe navigation, i.e. optical camera, LiDAR, GPS, Inertial Measurement Unit (IMU), and two computer vision (CV) detection models. All robots cooperate with a single RC, having its own CV model, and is capable of controlling the robots' motion along axes, rotation, and camera roll. Together, these agents are tasked with detecting a wildfire incident with the objective to meet a required detection confidence as fast as possible, while ensuring that the detection task is completed within a predetermined time limit.

Accordingly, we structure the evaluation around three key points: \textbf{1)} the trade-off between time and mission reliability, \textbf{2)} the importance of hierarchical coordination and robotic cooperation and \textbf{3)} the effectiveness of D$^3$ARC compared to non-cooperative and heuristic baselines. 

\textbf{Evaluation Metrics}: To quantitatively evaluate the performance of D$^3$ARC, Table \ref{tab:acronyms} summarizes the evaluated metrics. All metrics are averaged across missions and robots. The look-forward accuracy (LFA) is expressed as:
\begin{equation}
    \text{LFA} = \frac{\sum_{r \in \mathcal{R}}(1- \frac{MAE_{r}}{MAE^{\text{max}}_{r}})}{|\mathcal{R}|} \cdot 100\%, \notag
\end{equation}
where the $MAE_{r}$ expresses the mean absolute error between the anticipated and actual reward for robot $r$ in a single mission, and $MAE^{\text{max}}_{r}$ expresses the maximum MAE for robot $r$ across missions. 

\begin{table}[!ht]
\centering
\resizebox{\linewidth}{!}{%
\begin{tabular}{|
>{\columncolor[HTML]{EFEFEF}}p{3.1cm} |p{5.2cm}|}
\hline
\multicolumn{1}{|c|}{\cellcolor[HTML]{EFEFEF}\textbf{Metric}} & \multicolumn{1}{c|}{\cellcolor[HTML]{EFEFEF}\textbf{Definition}} \\ \hline
\cellcolor[HTML]{EFEFEF}\textbf{Total Time (min)} & Total mission time\\ \hline
\cellcolor[HTML]{EFEFEF}\textbf{Computation Time (min)} & Time for CV execution \\ \hline 
\cellcolor[HTML]{EFEFEF}\textbf{Transmission Time (min)} & Time during sensing data transmission \\ \hline 
\cellcolor[HTML]{EFEFEF}\textbf{Movement Time (min)} & Time taken during movement \\ \hline 
\cellcolor[HTML]{EFEFEF}\textbf{MTLR (\%)} & Mission Time Limit Reached \\ \hline
\textbf{TUFD (min)} & Time Until First valid Detection \\ \hline 
\textbf{MD$_{AVG}$ (min)} & Average Mission Duration \\ \hline 
\textbf{MD$_{Median}$ (min)} & Median Mission Duration \\ \hline 
\cellcolor[HTML]{EFEFEF}\textbf{MTP (\%)} & Movement Time Percentage \\ \hline
\cellcolor[HTML]{EFEFEF}\textbf{CTP (\%)} & Computation Time Percentage \\ \hline
\cellcolor[HTML]{EFEFEF}\textbf{TTP (\%)} & Transmission Time Percentage \\ \hline
\cellcolor[HTML]{EFEFEF}\textbf{FPDs (\%)} & False Positive Detections Rate \\ \hline
\cellcolor[HTML]{EFEFEF}\textbf{MSCS (\%)} & Stable Confidence Score over a mission\\ \hline
\cellcolor[HTML]{EFEFEF}\textbf{MLFA (\%)} & Look-Forward Accuracy \\ \hline
\textbf{MSR (\%)} & Mission Success Rate, excluding FPDs \\ \hline
\end{tabular}%
}
\caption{Evaluation Metrics Definition}
\label{tab:acronyms}
\end{table}
\vspace{-8pt}
\subsection{Simulation Setup}\label{simulation_setup}

This section outlines our simulation ecosystem, including: \textbf{1)} the 3D environment, \textbf{2)} the UAV robots, \textbf{3)} the CV detection models, and \textbf{4)} the wireless communication network.

\begin{figure}[!ht]
    \centering
    \includegraphics[width=0.68\linewidth]{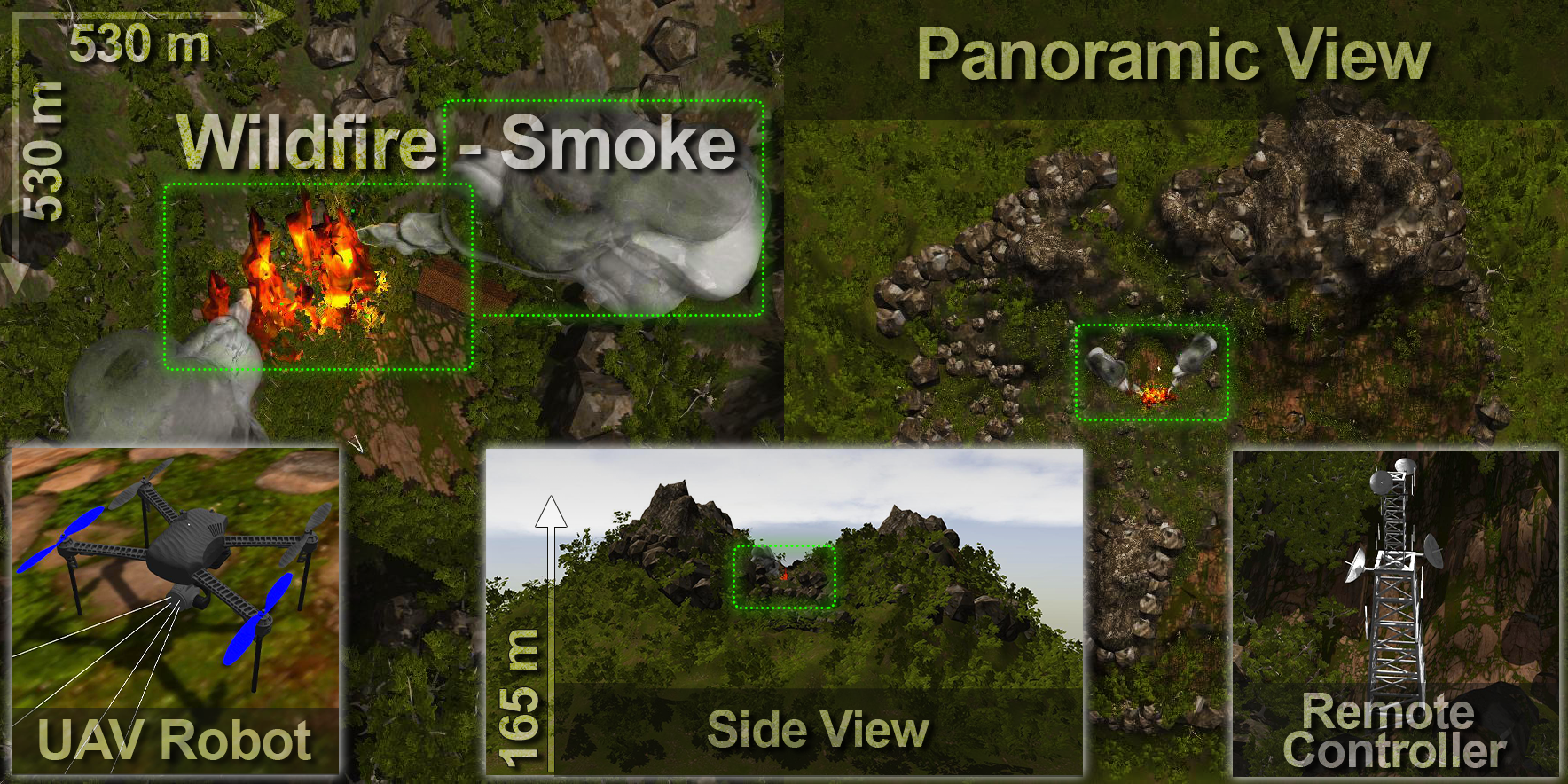}
    \caption{Simulation Environment inside Gazebo}
    \label{gazebo_environment}
\end{figure}

\textbf{3D Environment}: The simulation environment is built upon Gazebo 11 \cite{gazebo} and Robot Operating System 2 (ROS2) \cite{ros2}, incorporating rigid-body dynamics and collision physics through the Open Dynamics Engine \cite{ode}. These components provide a physics-aware simulation setting for evaluating robotic behavior under realistic terrain and environmental conditions. Figure \ref{gazebo_environment} illustrates the environment that includes a custom 530 x 530 x 165 m (width x depth x height) mountainous forest scene along with a 46 m high wildfire (including smoke), all designed in Blender \cite{blender}.

\textbf{UAV Robots}: The simulated UAV robots' properties were highly inspired by the 3DR Iris quadcopter to provide a realistic reference for the robot models \cite{iris_drone}. Table \ref{tab:robot_setup} given in the Appendix, summarizes the UAV robot configuration (same configuration is applied to all robots), including the set of sensors, hardware specifications, computational and communication capabilities, and key physical properties.

\textbf{CV Detection Models}: Two custom convolutional neural network (CNN) models are available for wildfire detection: a Full CNN and a Quantized CNN, differing in computational complexity. Both models are available for all UAV robots, while the $\text{RC}$ uses only the Full one. Details regarding their training and configuration are provided in the Appendix (Table \ref{tab:cv_models}).

\textbf{Wireless Communication Network}: The communication model considers a single Base Station (BS). The $\text{RC}$ is assumed to be collocated with the BS, and as such share the same physical location and jointly represent the communication endpoint of each UAV robot. A 2.4 GHz wireless link connecting each robot and the RC is characterized by distance-dependent path loss, shadow fading, bounded bandwidth, and spectral efficiency. The transmit power is determined with respect to a target received signal strength, while the complete network parameterization is provided in Appendix (Table \ref{tab:network_setup}).

\textbf{Ecosystem Dynamicity}:  
To simulate realistic conditions, introduce dynamicity and uncertainty into the environment and evaluate the generalization ability of D$^3$ARC, several key parameters are varied at each timestep. These variations capture changes in: \textbf{1)} robot's resource availability, \textbf{2)} wildfire locations and \textbf{3)} detection capabilities.  
\begin{itemize}
    \item \textbf{Resource availability}: The computational and communication resources vary across timesteps and robots (CPU speed, assigned bandwidth). This also applies to the wireless communication quality, including degradation with distance and fluctuations in the achievable data rate and transmission power requirements (Section \ref{problem_formulation}).
    \item \textbf{Wildfire locations}: The environment includes 14 distinct wildfire locations, with one wildfire incident per mission, and two possible BS locations. The distance between each wildfire location and the corresponding BS ranges from 18 to 275 m. The wildfire locations introduce detection difficulty (e.g. distant and obscured), while alternative BS placements impact connectivity and detection reliability.
    \item \textbf{Detection capabilities}: To emulate the heterogeneous perception capabilities introduced in Section \ref{system_model}, at each mission we randomly select a subset of robots (up to $50\%$ of the total robots) as \textit{limited-detection} UAVs and cap their detection confidence below the mission threshold ($\eta_{\min}$). The remaining robots operate as \textit{standard-detection} UAVs without this restriction. This prevents all robots from independently achieving reliable detections under ideal sensing conditions and promotes scenarios where successful mission completion depends on stronger coordination and cooperative sensing and detection.

\end{itemize}

\subsection{D$^3$ARC Configuration Setup}\label{configuration_setup}

Table \ref{tab:dsar_setup} summarizes the configuration of D$^3$ARC regarding the region size, motion command ranges, thresholds and key algorithmic parameters. Table \ref{tab:DSAR-NN-Configuration} summarizes the configuration for the RFNN models used by each UAV Robot and the $\text{RC}$. Overall, 115K samples, a batch size of 128, the Adam optimizer with learning rate 5e-06 and 2000 epochs were used for the offline training phase with 85\%, 10\%, and 5\% of the samples used for the training, validation and evaluation stages. To avoid overfitting, dropout layers were used with 10\% probability of zeroing neural connections along with an early stopping where the patience was set to 10 epochs. All hyperparameters are empirically tuned during the offline warmup phase based on mission success, detection reliability, and training stability. Once set, all parameter values remain fixed across the online evaluation missions.

\begin{table}[!ht]
\centering
\resizebox{220pt}{!}{%
\begin{tabular}{|
>{\columncolor[HTML]{EFEFEF}}l |l|l|}
\hline
\textbf{Entity RFNN} & \cellcolor[HTML]{EFEFEF}\textbf{\# Input Features} & \cellcolor[HTML]{EFEFEF}\textbf{\# RFNN Layers: {[}Shape{]}} \\ \hline
\textbf{UAV Robot} & 11 ($S_{k,r} \cup A_{k,r}$) & 3: {[}128 x 64 x 32{]} \\ \hline
\textbf{$\text{RC}$} & 13 ($S_{\text{RC},k,r} \cup A_{\text{RC},k,r}$) & 4: {[}128 x 64 x 64 x 32{]} \\ \hline
\end{tabular}%
}
\caption{D$^3$ARC: RFNNs for the UAV Robot \& $\text{RC}$}
\label{tab:DSAR-NN-Configuration}
\end{table}
\begin{table}[!ht]
\centering
\resizebox{220pt}{!}{%
\begin{tabular}{|
>{\columncolor[HTML]{EFEFEF}}l |l|}
\hline
\multicolumn{1}{|c|}{\cellcolor[HTML]{EFEFEF}\textbf{Parameter}} & \multicolumn{1}{c|}{\cellcolor[HTML]{EFEFEF}\textbf{Value}} \\ \hline
\textbf{Cubic Region Size} (\textit{crs}) & 60x60x60m \\ \hline
\textbf{Axes Movement Range}($\delta^{(x)},\delta^{(y)}, \delta^{(z)}$) & [-20,20]m \\ \hline
\textbf{Body Rotation Range} ($\delta^{(\mathtt{y})}$)& [-90,90]$\degree$  \\ \hline
\textbf{Camera Roll Range} ($\bm{\delta}^{\textbf{(so)}}$)& [0,90]$\degree$  \\ 
\hline
\textbf{Time tolerance} ($\mathtt{tl}$) & 200ms\\ \hline
\textbf{\# Candidate Actions} ($N$) & 1000 \\ \hline
\textbf{Penalty Weights} ($\lambda1,\lambda2, \lambda3$) & 1,1,1\\ \hline
\textbf{Objective Weights} ($\mu1,\mu2, \mu3, \mu4$) & 1,2,1,2 \\ \hline
\textbf{Minimum Confidence Threshold} ($\eta_{\min}$) & 75\% \\ \hline
\textbf{Intermediate Confidence Threshold} ($\eta_{int}$) & 20\% \\ \hline
\textbf{Decrease Factors} (\textbf{fv}) & [3,10,10]\% \\ \hline
\textbf{CFA Distance Radius} ($\delta_{CFA}$) & 100m\\ \hline
\textbf{CFA Factor} ($\mathtt{fv}_{CFA}$) & 2 \\ \hline
\textbf{Early Mission Completion Patience} ($esp$) & 2 \\ \hline
\textbf{Consecutive Low Confidence Threshold} ($c_{\mathrm{thr}}$) & 2 \\ \hline
\end{tabular}%
}
\caption{D$^3$ARC: Configuration Setup}
\label{tab:dsar_setup}
\end{table}

\subsection{Time-Aware Multi-Robot: Evaluation Results} \label{TA_Results}

This section presents the evaluation of D$^3$ARC under time-sensitive wildfire detection requirements. The evaluation follows a progressive analysis of 3 rounds, with 300 missions per setup, to examine the main factors that affect mission reliability, cooperative sensing, detection performance, and scalability. First, a single robot case is evaluated under different mission time limits to study whether increasing the available time is sufficient to achieve reliable detection. Then, an ablation study is conducted by varying the number of robots and comparing D$^3$ARC under cooperative and non cooperative configurations. This aims to show that the benefits do not come solely from increasing the number of robots, but also from targeted cooperation among them. Finally, D$^3$ARC is compared against a distributed heuristic cooperative strategy to evaluate its benefit over a baseline. 

\textbf{Results 1: Single Robot Time Limit Sensitivity}

This first round of experiments evaluates the performance of a single \textit{standard-detection} UAV robot under different mission time limits (3, 5, 10 min). The goal is to examine whether increasing the available mission time is sufficient to improve wildfire detection reliability, or whether the single robot setup remains limited by its own sensing and exploration capabilities.
\begin{figure}[!ht]
    \centering
    \includegraphics[width=1.0\linewidth]{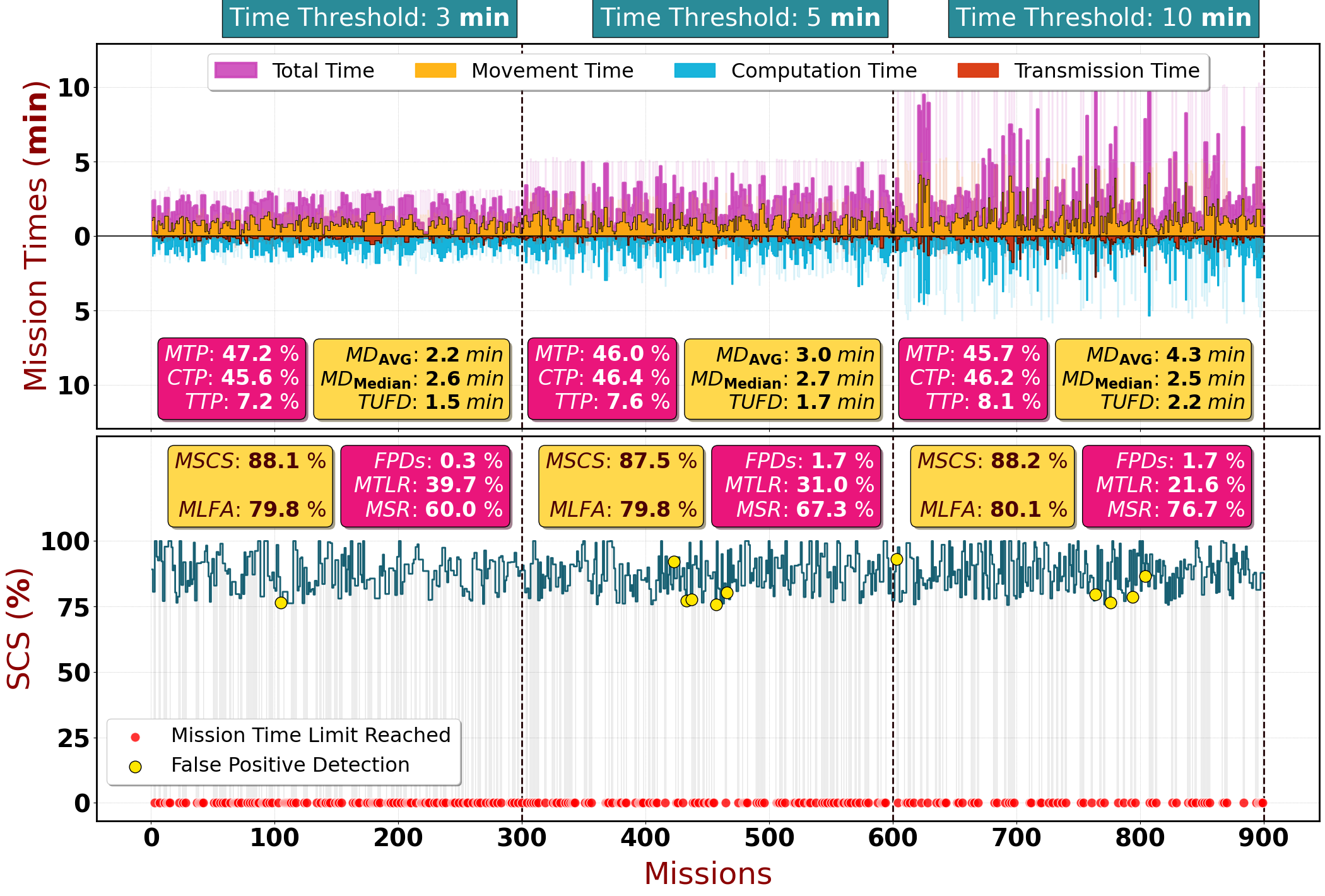}
    \caption{[\textbf{Single Robot}]: Performance for [3, 5, 10] minutes}
    \label{fig:mr_ta_reward_performance1}
\end{figure}

\textbf{Results 1 (Figure \ref{fig:mr_ta_reward_performance1}) - Observations:}
\begin{itemize}
\item The mean LF accuracy (MLFA) remains stable around 80\% across all thresholds, indicating high decision-making quality for both agents.
\item The average mission duration ($\text{MD}_{\text{avg}}$) and time until first detection (TUFD) remain on average at $\sim$48\% and $\sim$35\% of their respective thresholds, indicating that most missions are completed early. Higher time thresholds increase mission duration as they allow further exploration of harder cases.
\item Movement and computation dominate the total mission time, with each accounting for $\sim$46\% of the mission duration, while transmission remains below $\sim$8\%. 
\item As time threshold increases,  the percentage of missions reaching the time limit (MTLR) decreases. This indicates that the robot is allowed to search more regions and increase the probability to meet the detection performance requirement.
\item However, increasing the time threshold by more than three times, reaching the maximum acceptable by EUSPA (10 min) \cite{EUSPA2024EMAidUserNeeds}, the mission success ratio (MSR) still remains under 77\%, although detection performance remains high ($\uparrow$ MSCS, $\downarrow$ FPDs) and consistent across all time limits. 
\end{itemize}

\textbf{Results 1 - Conclusion:} The single robot results show that increasing the time limit improves overall performance. However, all remarks reveal an important limitation of the single robot setup. This limitation is not the quality of the detection once the robot reaches a promising region, but the robot’s ability to consistently reach such regions within the required time. In simple words, \textit{the robot can detect reliably, yet time is not enough}. This motivates the transition to a multi-robot setup, where cooperation can provide additional sensing perspectives, faster coverage, and stronger support for uncertain detections.

\textbf{Results 2: Impact of Multi-Robot Cooperation under a Strict Time Limit}

Transitioning into a multi-robot setup, this round of results quantifies the impact of cooperation under the strictest mission time limit of 3 min, and for up to 4 UAV robots. The single robot is included for reference. The goal is to examine whether increasing only the number of robots improves mission reliability, or explicit cooperation is required. To compare D$^3$ARC with and without cooperation, we enable/disable the two cooperative mechanisms, i.e. the coverage efficiency and call-for-aid (Section \ref{proposed_algorithm}). In each multi-robot case, 50\% of the robots are configured as \textit{limited-detection} UAVs. 

\begin{figure*}[!ht]
    \centering

    \begin{subfigure}{0.68\linewidth}
        \centering
        \includegraphics[width=\linewidth]{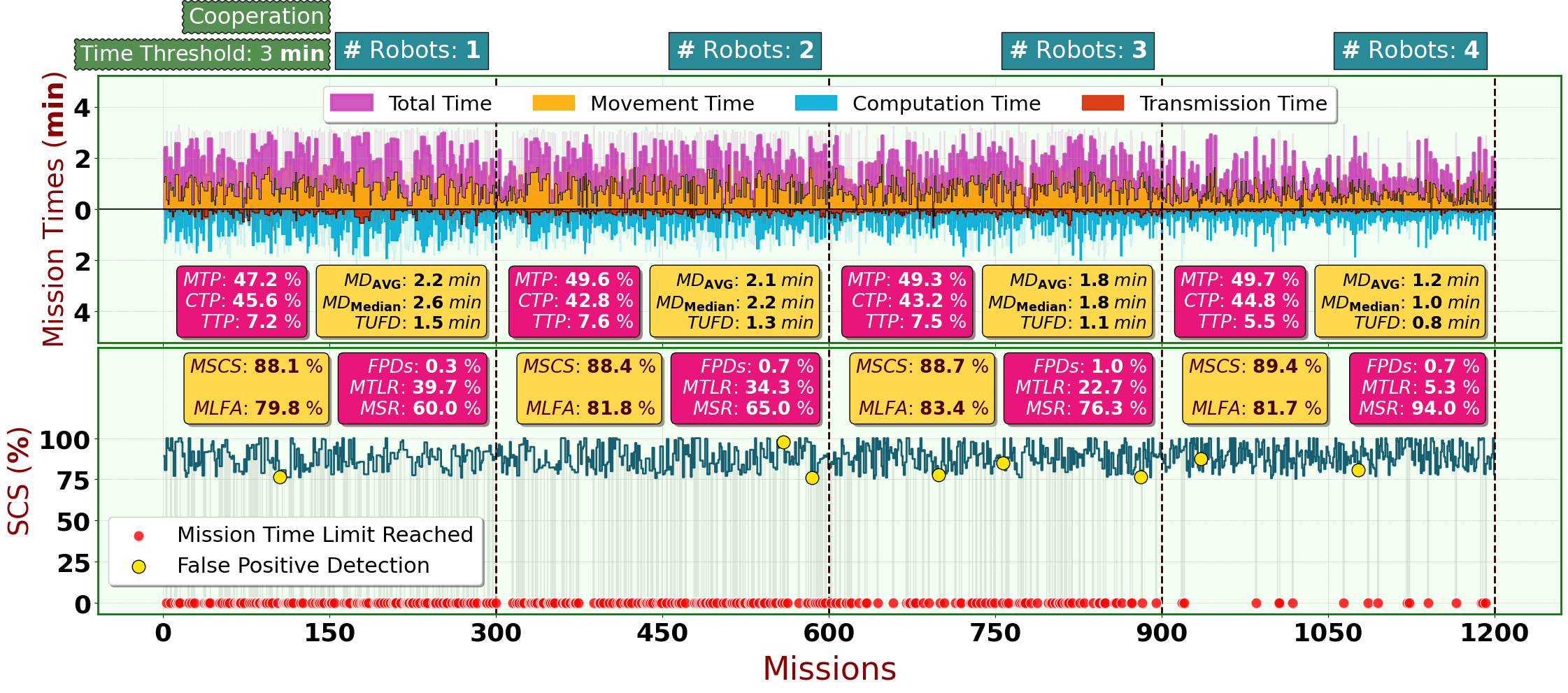}
        \caption{Cooperative Setup}
        \label{fig:mr_ta_reward_performance}
    \end{subfigure}
    \begin{subfigure}{0.68\linewidth}
        \centering
        \includegraphics[width=\linewidth]{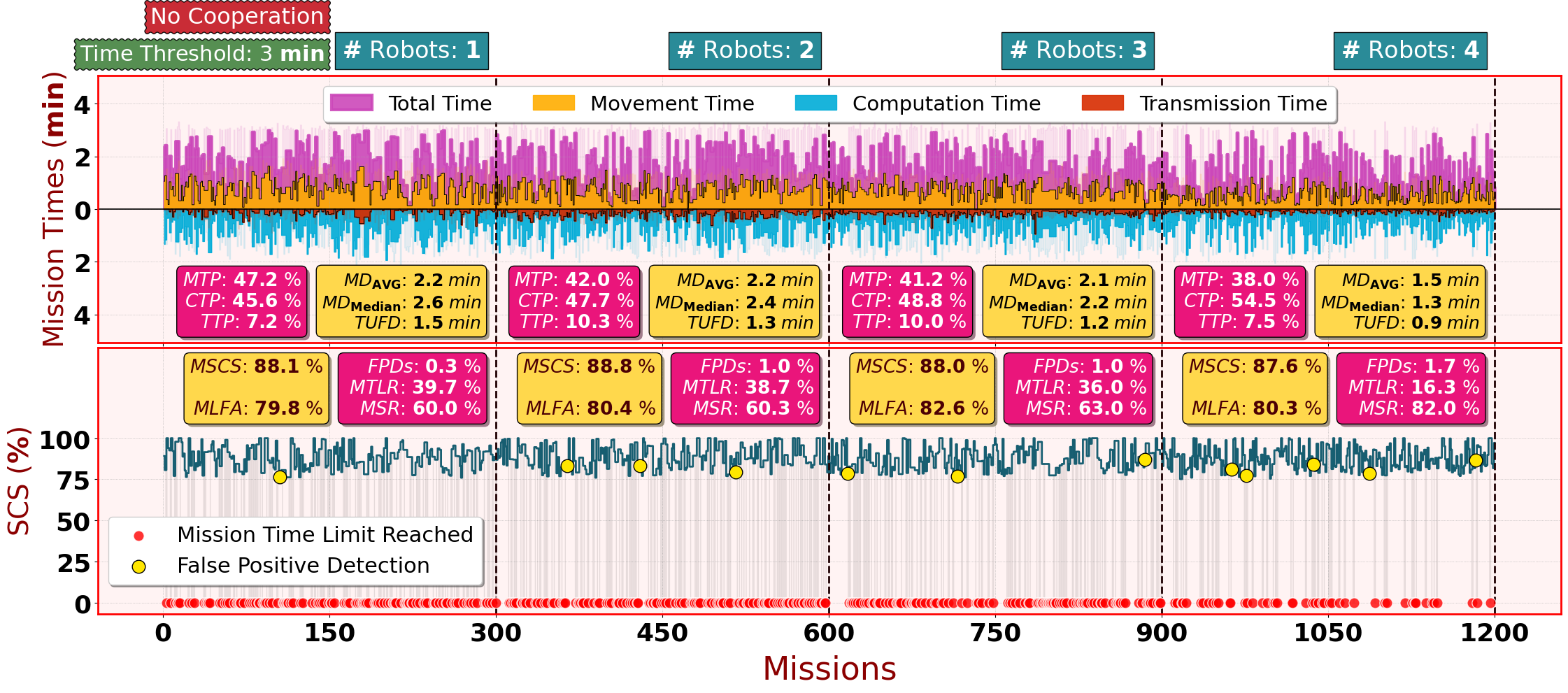}
        \caption{Non-Cooperative Setup}
        \label{fig:mr_ta_time_performance}
    \end{subfigure}

    \caption{[\textbf{Multi-Robot}]: Performance for the strictest time threshold of 3 minutes.}
    \label{fig:mr_ta_performance}
\end{figure*}

\textbf{Results 2 (Figure \ref{fig:mr_ta_performance}) - Observations:} 
\begin{itemize} 
\item Looking at the MSR, the benefit of increasing the number of robots becomes evident. Compared with the single robot case, cooperation improves the MSR by 5\%, 16.3\%, and 34\%, for 2, 3, and 4 robots, respectively, while without cooperation, by 0.3\%, 3\%, and 22\%. 
\item Cooperation provides an additional gain in the MSR of 4.7\%, 13.3\%, and 12\% over the non-cooperative setup. This highlights that even when multiple robots are available, coordinated support is necessary to use the robots' capabilities effectively and avoid relying on independent detections alone.
\item Cooperation also reduces time limit violations, decreasing the MTLR by 4.4\%, 13.3\%, and 11\% as robots increase, compared with the non-cooperative setup. This shows that cooperation does not only improve mission success rates, but facilitates robots to satisfy the detection objective earlier and more reliably.
\item Cooperation reduces $\text{MD}_{\text{avg}}$ as the number of robots increases, with more evident reductions of 14\% and 20\% for 3 and 4 robots, respectively. It also significantly lowers TUFD, reaching 0.8 min with 4 robots, showing that cooperation guides robots toward promising regions faster.
\item The cooperative setup maintains high MSCS and low FPDs, reaching 89.4\% and 0.7\%, respectively, with 4 robots, compared with 87.6\% and 1.7\% without cooperation. This shows that cooperation accelerates mission completion while also preserving stronger detection confidence and almost eliminating incorrect wildfire detections. 
\item The MLFA improves under cooperation, indicating that cooperative information facilitate the agents selecting more effective actions toward promising regions. 
\item The time distribution also changes with cooperation. Cooperative configurations spend a higher amount of time on movement, especially in the case of 4 robots, where MTP is 23\% higher, while the non-cooperative setup spends more time on computation. This is mainly a result of D$^3$ARC's coverage efficiency mechanism, which discourages repeated visits to \textit{uninformative} regions and incentivizes the robots to continue exploring the environment. 
\end{itemize} 

\textbf{Results 2 - Conclusion:} The results show that cooperation can convert additional robots into effective sensing support under strict time constraints. This results in higher mission success, fewer time limit violations, shorter mission duration, higher coverage, and earlier detections compared to the non-cooperative configuration. In simple words, \textit{multiple robots may travel far, yet cooperative ones arrive with purpose}.

\textbf{Results 3: Baseline Comparison}

This final round of evaluation results compares the cooperative D$^3$ARC and non-cooperative (marked as D$^3$AR) against CH-RBS, a Rule-Based Strategy (see Appendix \ref{appendix}) that integrates the cooperative mechanisms of D$^3$ARC. The D$^3$AR is included just for reference. This time we examine the merits of adaptive intelligence realized through the hierarchical distributed agentic decision-making of D$^3$ARC. The comparison is conducted using the 3 robot setup under the strict 3 min time threshold. This setting represents a challenging middle-ground scenario that maintains coordination and cooperation complexity.
\begin{table}[!ht]
\centering
\resizebox{250pt}{!}{%
\begin{tabular}{|ccccc|}
\hline
\rowcolor[HTML]{656565} 
\multicolumn{5}{|c|}{\cellcolor[HTML]{656565}{\color[HTML]{FFFFFF} \textbf{\# Robots: 3 - Time Threshold: 3 minutes}}} \\ \hline
\rowcolor[HTML]{EFEFEF} 
\multicolumn{1}{|c|}{\cellcolor[HTML]{EFEFEF}\textbf{Strategy}} & \multicolumn{1}{c|}{\cellcolor[HTML]{EFEFEF}\textbf{\begin{tabular}[c]{@{}c@{}}Avg. Mission Duration\\ (min)\end{tabular}}} & \multicolumn{1}{c|}{\cellcolor[HTML]{EFEFEF}\textbf{\begin{tabular}[c]{@{}c@{}}Avg. TUFD\\ (min)\end{tabular}}} & \multicolumn{1}{c|}{\cellcolor[HTML]{EFEFEF}\textbf{\begin{tabular}[c]{@{}c@{}}MSCS\\ (\%)\end{tabular}}} & \cellcolor[HTML]{EFEFEF}\textbf{\begin{tabular}[c]{@{}c@{}}Avg. MSR\\ (\%)\end{tabular}} \\ \hline
\rowcolor[HTML]{FFFFFF} 
\multicolumn{1}{|c|}{\cellcolor[HTML]{EFEFEF}\textbf{D$^3$ARC}} & \multicolumn{1}{c|}{\cellcolor[HTML]{FFFFFF}\textbf{1.8}} & \multicolumn{1}{c|}{\cellcolor[HTML]{FFFFFF}\textbf{1.1}} & \multicolumn{1}{c|}{\cellcolor[HTML]{FFFFFF}\textbf{88.7}} & \textbf{76.3} \\ \hline
\rowcolor[HTML]{FFFFFF} 
\multicolumn{1}{|c|}{\cellcolor[HTML]{EFEFEF}\textbf{D$^3$AR}} & \multicolumn{1}{c|}{\cellcolor[HTML]{FFFFFF}2.1} & \multicolumn{1}{c|}{\cellcolor[HTML]{FFFFFF}1.2} & \multicolumn{1}{c|}{\cellcolor[HTML]{FFFFFF}88.0} & 63.0 \\ \hline
\rowcolor[HTML]{FFFFFF} 
\multicolumn{1}{|c|}{\cellcolor[HTML]{EFEFEF}\textbf{CH-RBS}} & \multicolumn{1}{c|}{\cellcolor[HTML]{FFFFFF}2.3} & \multicolumn{1}{c|}{\cellcolor[HTML]{FFFFFF}1.6} & \multicolumn{1}{c|}{\cellcolor[HTML]{FFFFFF}87.5} & 44.0 \\ \hline
\end{tabular}%
}
\caption{Comparison of D$^3$ARC against Baselines}
\label{tab:baseline_comparisons_TA}
\end{table}

\textbf{Results 3 (Table \ref{tab:baseline_comparisons_TA}) - Observations:} 
\begin{itemize} 
\item D$^3$ARC achieves the highest MSR, reaching 76.3\%, compared to 44\% of CH-RBS. This large gap indicates that heuristic-driven cooperation rules are not sufficient under dynamic ecosystem conditions. D$^3$ARC adapts agentic decision-making, allowing the robots to use cooperation more effectively. 
\item D$^3$ARC achieves the shortest average mission duration, reducing it from 2.3 min in CH-RBS to 1.8 min. This shows that D$^3$ARC does not only increase mission success, but also enables faster completion under the same time threshold. 
\item The average TUFD is also 32\% lower compared to CH-RBS. This indicates that D$^3$ARC supports robots in detecting informative wildfire evidence earlier in the mission. 
\item Although the differences in MSCS are moderate, they show that the faster mission completion and highest MSR of D$^3$ARC is not achieved at the cost of detection confidence. 
\end{itemize} 

\textbf{Results 3 - Conclusion:} The results confirm that by combining multiple robots, cooperative sensing and hierarchical distributed agentic decision-making, D$^3$ARC dominates. In simple words, \textit{cooperation matters, yet adaptive intelligence makes the difference}.

\section{Conclusions} \label{conclusions}
This paper introduces D$^3$ARC, a time-aware distributed, hierarchical, and cooperative framework integrating multiple robots and a remote controller for wildfire detection under \textit{uncertainty}. The controller decides each robot’s motion, while each robot determines \textit{where} and \textit{how} to perform detection. All agents aim to detect wildfire with a certain performance as fast as possible under an imposed time limit. D$^3$ARC integrates an innovative forward-looking approach, along with four key mechanisms that support safe operation while promoting exploration, cooperation and system efficiency.

The results demonstrate the effectiveness of D$^3$ARC. Single robot experiments show reliable detections, but with a time cost. Multi-robot cooperation improves coverage speed and mission duration, enables earlier and reliable detections, and achieves mission success up to 94\% with 89.4\% detection confidence. Comparisons with the non-cooperative D$^3$AR and the cooperative heuristic CH-RBS confirm the merits of D$^3$ARC’s adaptive, confidence-driven, and time-aware distributed yet cooperative decision-making.

\vspace{-4pt}
\bibliographystyle{IEEEtran}
\bibliography{bibliography/bib}

\begin{thebibliography}{10}
\providecommand{\url}[1]{#1}
\csname url@samestyle\endcsname
\providecommand{\newblock}{\relax}
\providecommand{\bibinfo}[2]{#2}
\providecommand{\BIBentrySTDinterwordspacing}{\spaceskip=0pt\relax}
\providecommand{\BIBentryALTinterwordstretchfactor}{4}
\providecommand{\BIBentryALTinterwordspacing}{\spaceskip=\fontdimen2\font plus
\BIBentryALTinterwordstretchfactor\fontdimen3\font minus \fontdimen4\font\relax}
\providecommand{\BIBforeignlanguage}[2]{{%
\expandafter\ifx\csname l@#1\endcsname\relax
\typeout{** WARNING: IEEEtran.bst: No hyphenation pattern has been}%
\typeout{** loaded for the language `#1'. Using the pattern for}%
\typeout{** the default language instead.}%
\else
\language=\csname l@#1\endcsname
\fi
#2}}
\providecommand{\BIBdecl}{\relax}
\BIBdecl

\bibitem{wmo2026globalclimate}
\BIBentryALTinterwordspacing
{World Meteorological Organization}, ``{State of the Global Climate 2025},'' World Meteorological Organization, Geneva, Switzerland, Tech. Rep. WMO-No. 1391, Mar. 2026. [Online]. Available: \url{https://library.wmo.int/records/item/69807-state-of-the-global-climate-2025}
\BIBentrySTDinterwordspacing

\bibitem{EUSPA2024EMAidUserNeeds}
\BIBentryALTinterwordspacing
``User needs and requirements: Report on emergency management and humanitarian aid,'' European Union Agency for the Space Programme (EUSPA), Tech. Rep., 2024. [Online]. Available: \url{www.euspa.europa.eu/publications-multimedia/publications/user-needs-and-requirements}
\BIBentrySTDinterwordspacing

\bibitem{Ghassemian2026}
M.~Ghassemian, J.~Erfanian, K.~Althoefer, K.~Ahmadi, J.~Eichinger, A.~Meseguer~Valenzuela, P.~Barattini, M.~Bahaei, T.~Booth, A.~Bouttier, C.~Ciochina, M.~Mestoukirdi, P.~Gonçalves, E.~Natalizio, J.~Simonjan, M.~Basaran, F.~Cogen, Y.~Ozsahin, A.~Huete, and T.~Mahmoodi, ``6g architectural foundations and ai-native solutions for future connected robotics.'' 03 2026.

\bibitem{6G-IA_2024_EuropeanVision}
\BIBentryALTinterwordspacing
{6G-IA Vision Working Group}, ``European vision for the 6g network ecosystem,'' 6G-IA, Tech. Rep. Version 2.0, 2024. [Online]. Available: \url{https://6g-ia.eu/wp-content/uploads/2024/11/european-vision-for-the-6g-network-ecosystem.pdf}
\BIBentrySTDinterwordspacing

\bibitem{XIONG2025100915}
\BIBentryALTinterwordspacing
Y.~Xiong, Y.~Zhou, J.~She, and A.~Yu, ``Collaborative coverage path planning for uav swarm for multi-region post-disaster assessment,'' \emph{Vehicular Communications}, vol.~53, p. 100915, 2025. [Online]. Available: \url{https://www.sciencedirect.com/science/article/pii/S2214209625000427}
\BIBentrySTDinterwordspacing

\bibitem{Aminzadeh2023}
\BIBentryALTinterwordspacing
A.~Aminzadeh and A.~M. Khoshnood, ``Multi-uav cooperative search and coverage control in post-disaster assessment: Experimental implementation,'' \emph{Intelligent Service Robotics}, vol.~16, no.~4, pp. 415--430, 2023. [Online]. Available: \url{https://doi.org/10.1007/s11370-023-00476-4}
\BIBentrySTDinterwordspacing

\bibitem{Farsath10580372}
K.~Rashida~Farsath, K.~Jitha, V.~Mohammed~Marwan, A.~Muhammed Ali~Jouhar, K.~Muhammed~Farseen, and K.~Musrifa, ``Ai-enhanced unmanned aerial vehicles for search and rescue operations,'' in \emph{2024 5th International Conference on Innovative Trends in Information Technology (ICITIIT)}, 2024, pp. 1--10.

\bibitem{XING2022102972}
\BIBentryALTinterwordspacing
L.~Xing, X.~Fan, Y.~Dong, Z.~Xiong, L.~Xing, Y.~Yang, H.~Bai, and C.~Zhou, ``Multi-uav cooperative system for search and rescue based on yolov5,'' \emph{International Journal of Disaster Risk Reduction}, vol.~76, p. 102972, 2022. [Online]. Available: \url{https://www.sciencedirect.com/science/article/pii/S2212420922001911}
\BIBentrySTDinterwordspacing

\bibitem{Horyna101007}
\BIBentryALTinterwordspacing
J.~Horyna, T.~Baca, V.~Walter, D.~Albani, D.~Hert, E.~Ferrante, and M.~Saska, ``Decentralized swarms of unmanned aerial vehicles for search and rescue operations without explicit communication,'' \emph{Auton. Robots}, vol.~47, no.~1, p. 77–93, Oct. 2022. [Online]. Available: \url{https://doi.org/10.1007/s10514-022-10066-5}
\BIBentrySTDinterwordspacing

\bibitem{romero2024}
\BIBentryALTinterwordspacing
A.~Romero, C.~Delgado, L.~Zanzi, R.~Suárez, and X.~Costa-Pérez, ``Cellular-enabled collaborative robots planning and operations for search-and-rescue scenarios,'' 2024. [Online]. Available: \url{https://arxiv.org/abs/2403.09177}
\BIBentrySTDinterwordspacing

\bibitem{Medeiros9609965}
I.~Medeiros, A.~Boukerche, and E.~Cerqueira, ``Swarm-based and energy-aware unmanned aerial vehicle system for video delivery of mobile objects,'' \emph{IEEE Transactions on Vehicular Technology}, vol.~71, no.~1, pp. 766--779, 2022.

\bibitem{BECK2018251}
\BIBentryALTinterwordspacing
Z.~Beck, W.~L. Teacy, A.~Rogers, and N.~R. Jennings, ``Collaborative online planning for automated victim search in disaster response,'' \emph{Robotics and Autonomous Systems}, vol. 100, pp. 251--266, 2018. [Online]. Available: \url{https://www.sciencedirect.com/science/article/pii/S0921889016307515}
\BIBentrySTDinterwordspacing

\bibitem{GHASSEMI2022103905}
\BIBentryALTinterwordspacing
P.~Ghassemi and S.~Chowdhury, ``Multi-robot task allocation in disaster response: Addressing dynamic tasks with deadlines and robots with range and payload constraints,'' \emph{Robotics and Autonomous Systems}, vol. 147, p. 103905, 2022. [Online]. Available: \url{https://www.sciencedirect.com/science/article/pii/S0921889021001901}
\BIBentrySTDinterwordspacing

\bibitem{khanal2025}
\BIBentryALTinterwordspacing
A.~Khanal, J.~P. Mathew, C.~Nowzari, and G.~J. Stein, ``Learning-augmented model-based multi-robot planning for time-critical search and inspection under uncertainty,'' 2025. [Online]. Available: \url{https://arxiv.org/abs/2507.06129}
\BIBentrySTDinterwordspacing

\bibitem{Han8040138}
\BIBentryALTinterwordspacing
D.~Han, H.~Jiang, L.~Wang, X.~Zhu, Y.~Chen, and Q.~Yu, ``Collaborative task allocation and optimization solution for unmanned aerial vehicles in search and rescue,'' \emph{Drones}, vol.~8, no.~4, 2024. [Online]. Available: \url{https://www.mdpi.com/2504-446X/8/4/138}
\BIBentrySTDinterwordspacing

\bibitem{9762674Yin}
R.~Yin, Y.~Shen, H.~Zhu, X.~Chen, and C.~Wu, ``Time-critical tasks implementation in mec based multi-robot cooperation systems,'' \emph{China Communications}, vol.~19, no.~4, pp. 199--215, 2022.

\bibitem{Lu8100564}
\BIBentryALTinterwordspacing
Y.~Lu, C.~Xu, and Y.~Wang, ``Joint computation offloading and trajectory optimization for edge computing uav: A knn-ddpg algorithm,'' \emph{Drones}, vol.~8, no.~10, 2024. [Online]. Available: \url{https://www.mdpi.com/2504-446X/8/10/564}
\BIBentrySTDinterwordspacing

\bibitem{8331947Pham}
H.~X. Pham, H.~M. La, D.~Feil-Seifer, and M.~C. Deans, ``A distributed control framework of multiple unmanned aerial vehicles for dynamic wildfire tracking,'' \emph{IEEE Transactions on Systems, Man, and Cybernetics: Systems}, vol.~50, no.~4, pp. 1537--1548, 2020.

\bibitem{9504947Shrestha}
K.~Shrestha, R.~Dubey, A.~Singandhupe, S.~Louis, and H.~La, ``Multi objective uav network deployment for dynamic fire coverage,'' in \emph{2021 IEEE Congress on Evolutionary Computation (CEC)}, 2021, pp. 1280--1287.

\bibitem{Patrinopoulou2024}
N.~Patrinopoulou, I.~Daramouskas, D.~Meimetis, V.~Lappas, and V.~Kostopoulos, ``A distributed framework for persistent wildfire monitoring with fixed wing uavs,'' \emph{Drones and Autonomous Vehicles}, vol.~1, no.~3, p. 10009, 2024.

\bibitem{Julian2019}
\BIBentryALTinterwordspacing
K.~D. Julian and M.~J. Kochenderfer, ``Distributed wildfire surveillance with autonomous aircraft using deep reinforcement learning,'' 2018. [Online]. Available: \url{https://arxiv.org/abs/1810.04244}
\BIBentrySTDinterwordspacing

\bibitem{Seraj2022}
\BIBentryALTinterwordspacing
E.~Seraj, A.~Silva, and M.~Gombolay, ``Multi-uav planning for cooperative wildfire coverage and tracking with quality-of-service guarantees,'' \emph{Autonomous Agents and Multi-Agent Systems}, vol.~36, no.~2, p.~39, 2022. [Online]. Available: \url{https://doi.org/10.1007/s10458-022-09566-6}
\BIBentrySTDinterwordspacing

\bibitem{11315192Akpomedaye}
B.~Akpomedaye, A.~Shalan, N.~A. Walee, and M.~Rahman, ``Energy-efficient uav surveillance for early wildfire detection using ai-driven image analysis,'' in \emph{2025 IEEE/ACS 22nd International Conference on Computer Systems and Applications (AICCSA)}, 2025, pp. 1--7.

\bibitem{10206033Suo}
J.~Suo, X.~Zhang, W.~Shi, and W.~Zhou, ``E3-uav: An edge-based energy-efficient object detection system for unmanned aerial vehicles,'' \emph{IEEE Internet of Things Journal}, vol.~11, no.~3, pp. 4398--4413, 2024.

\bibitem{Luo2024}
\BIBentryALTinterwordspacing
X.~Luo, A.~Rechardt, G.~Sun, K.~K. Nejad, F.~Yáñez, B.~Yilmaz, K.~Lee, A.~O. Cohen, V.~Borghesani, A.~Pashkov, D.~Marinazzo, J.~Nicholas, A.~Salatiello, I.~Sucholutsky, P.~Minervini, S.~Razavi, R.~Rocca, E.~Yusifov, T.~Okalova, N.~Gu, M.~Ferianc, M.~Khona, K.~R. Patil, P.-S. Lee, R.~Mata, N.~E. Myers, J.~K. Bizley, S.~Musslick, I.~P. Bilgin, G.~Niso, J.~M. Ales, M.~Gaebler, N.~A.~R. Murty, L.~Loued-Khenissi, A.~Behler, C.~M. Hall, J.~Dafflon, S.~D. Bao, and B.~C. Love, ``Large language models surpass human experts in predicting neuroscience results,'' \emph{Nature Human Behaviour}, 2024. [Online]. Available: \url{https://doi.org/10.1038/s41562-024-02046-9}
\BIBentrySTDinterwordspacing

\bibitem{gazebo}
\BIBentryALTinterwordspacing
Official gazebo website. [Online]. Available: \url{https://gazebosim.org}
\BIBentrySTDinterwordspacing

\bibitem{ros2}
\BIBentryALTinterwordspacing
Official ros website. [Online]. Available: \url{https://www.ros.org/}
\BIBentrySTDinterwordspacing

\bibitem{ode}
\BIBentryALTinterwordspacing
Open dynamics engine. [Online]. Available: \url{https://www.ode.org/}
\BIBentrySTDinterwordspacing

\bibitem{blender}
\BIBentryALTinterwordspacing
Blender free \& open source 3d software. [Online]. Available: \url{https://www.blender.org/download/}
\BIBentrySTDinterwordspacing

\bibitem{iris_drone}
\BIBentryALTinterwordspacing
3dr iris quadcopter specifications. [Online]. Available: \url{https://www.arducopter.co.uk/iris-quadcopter-uav.html}
\BIBentrySTDinterwordspacing

\bibitem{dfire_git}
\BIBentryALTinterwordspacing
Dfire dataset official github. [Online]. Available: \url{https://github.com/gaiasd/DFireDataset}
\BIBentrySTDinterwordspacing

\bibitem{dfire}
\BIBentryALTinterwordspacing
P.~V. A.~B. de~Venâncio, R.~J. Campos, T.~M. Rezende, A.~C. Lisboa, and A.~V. Barbosa, ``A hybrid method for fire detection based on spatial and temporal patterns,'' \emph{Neural Computing and Applications}, vol.~35, no.~13, pp. 9349--9361, 2023. [Online]. Available: \url{https://doi.org/10.1007/s00521-023-08260-2}
\BIBentrySTDinterwordspacing

\bibitem{m4sfwd_git}
\BIBentryALTinterwordspacing
M4sfwd dataset official github. [Online]. Available: \url{https://github.com/Philharmy-Wang/M4SFWD}
\BIBentrySTDinterwordspacing

\bibitem{m4sfwd}
\BIBentryALTinterwordspacing
G.~Wang, ``Multiple scenarios, multiple weather conditions, multiple lighting conditions and multiple wildfire objects synthetic forest wildfire dataset (m4sfwd),'' 2024. [Online]. Available: \url{https://dx.doi.org/10.21227/m9kz-bw61}
\BIBentrySTDinterwordspacing

\end{thebibliography}

\appendix \label{appendix}
\subsection{Achievable Data Rate Modeling}
The achievable data rate $\mathtt{d}_{k,r}$ of each robot $r \in \mathcal{R}^{ready}_k$  is given by:
\vspace{-4pt}
\begin{equation}
  \mathtt{d}_{k,r} = {b_{k,r} \cdot log_2\left(1 + \frac{g_{k,r} \cdot p_{k,r}}{b_{k,r} \cdot N_0}\right)},
\end{equation}
where: 
\begin{equation}
  g_{k,r} = 10 ^{-\frac{\mathtt{L}_0 + 10 \cdot \mathtt{e} \cdot \log_{10}{(\frac{\delta_{k,\text{RC},r}}{\mathtt{\delta_0}}})}{10}},
\end{equation}
\vspace{-8pt}
\begin{equation}
    \mathtt{L}_0 = 20 \cdot \log_{10}(\mathtt{\delta}_0) + 20 \cdot \log_{10}(\mathtt{f}_{k})  +20 \cdot \log_{10}(\frac{4 \cdot \pi}{\mathtt{c}}),
\end{equation}
\vspace{-8pt}
\begin{equation}
  p_{k} = tp_{k} + \mathtt{L_0}.
\end{equation}
$tp_{k}$ is the target signal strength ($dBm$) $\mathtt{L}_0$ is the path loss ($dB$) at the reference distance $\mathtt{\delta}_0$ ($meters$), $\delta_{k,\text{RC},r}$ is the distance between the robot and the $\text{RC}$ at $k$ ($meters$), $\mathtt{f}_{k}$ denotes the transmission carrier frequency ($Hz$), $\mathtt{e}$ is the path loss exponent and $\mathtt{c}$ the speed of light ($m/s$).

\subsection{UAV Robot Specifications}

Table \ref{tab:robot_setup} summarizes the UAV robot's configuration. Several abbreviations are used in the table: Frames Per Second (FPS), Width ($\times$) Height (WxH), and Height/Width/Depth (H/W/D).

\begin{table}[h]
\centering
\begin{subtable}[t]{0.40\textwidth}
\centering
\resizebox{200pt}{!}{%
\begin{tabularx}{\linewidth}{
    |>{\columncolor[HTML]{EFEFEF}\raggedright\arraybackslash}X
    |>{\raggedright\arraybackslash}p{0.28\linewidth}|}
\hline
\multicolumn{1}{|c|}{\cellcolor[HTML]{EFEFEF}\textbf{Parameter}} &
\multicolumn{1}{c|}{\cellcolor[HTML]{EFEFEF}\textbf{Value}} \\ \hline
\textbf{\# of LiDARs} & 6 \\ \hline
\textbf{LiDAR Safety Distance} ($\mathtt{sd}_{\min}$) & 2 m \\ \hline
\textbf{Camera FPS} ($I_{\text{camera}}$) & 11 \\ \hline
\textbf{Camera Resolution (W$\times$H)} & $1920\times1080$ \\ \hline
\textbf{Camera Frame Size} ($w_{\text{camera}}$) & 519 KByte \\ \hline
\textbf{Sensing Period} ($\mathtt{s}$) & 3 s \\ \hline
\end{tabularx}
}
\caption{Sensing \& Perception}
\label{tab:robot_sensing}
\end{subtable}
\hfill
\begin{subtable}[t]{0.4\textwidth}
\centering
\resizebox{200pt}{!}{%
\begin{tabularx}{\linewidth}{
    |>{\columncolor[HTML]{EFEFEF}\raggedright\arraybackslash}X
    |>{\raggedright\arraybackslash}p{0.28\linewidth}|}
\hline
\multicolumn{1}{|c|}{\cellcolor[HTML]{EFEFEF}\textbf{Parameter}} &
\multicolumn{1}{c|}{\cellcolor[HTML]{EFEFEF}\textbf{Value}} \\ \hline
\textbf{CPU Speed} ($f^{\text{min}}/f^{\text{max}}$) & 1.5/3.2 GHz \\ \hline
\textbf{CPU Cores} ($n^{cpu}$) & 6 \\ \hline
\textbf{CPU ESC} ($\varsigma$) & $10^{-28}$ \\ \hline
\textbf{CPU FLOPs/Cycle} ($c$) & 8 \\ \hline
\textbf{Transmission Power} ($p^{\text{min}}/p^{\text{max}}$) & 14/33 dBm \\ \hline
\end{tabularx}
}
\caption{Computation \& Communication}
\label{tab:robot_compute}
\end{subtable}
\hfill
\begin{subtable}[t]{0.4\textwidth}
\centering
\resizebox{200pt}{!}{%
\begin{tabularx}{\linewidth}{
    |>{\columncolor[HTML]{EFEFEF}\raggedright\arraybackslash}X
    |>{\raggedright\arraybackslash}p{0.28\linewidth}|}
\hline
\multicolumn{1}{|c|}{\cellcolor[HTML]{EFEFEF}\textbf{Parameter}} &
\multicolumn{1}{c|}{\cellcolor[HTML]{EFEFEF}\textbf{Value}} \\ \hline
\textbf{Mass} & 1.816 kg \\ \hline
\textbf{\# of Propellers} & 4 \\ \hline
\textbf{Propeller Radius} & 100 mm \\ \hline
\textbf{Speed} ($v$) & 5 m/s \\ \hline
\end{tabularx}
}
\caption{Physical Specifications}
\label{tab:robot_physical}
\end{subtable}
\caption{UAV Robot Configuration Setup}
\label{tab:robot_setup}
\vspace{-10pt}
\end{table}

\subsection{Computer Vision Detection Model Configuration}

The Full and Quantized CV models were trained offline on publicly available datasets, namely DFire \cite{dfire_git,dfire} and M$^4$SFWD \cite{m4sfwd_git, m4sfwd}, which together contain approximately 25K diverse images. Their configurations are summarized in Table \ref{tab:cv_models}.

\begin{table}[!ht]
\centering
\begin{tabular}{|p{0.34\linewidth}|p{0.24\linewidth}|p{0.24\linewidth}|}
\hline
\cellcolor[HTML]{EFEFEF}\textbf{Configuration} & \cellcolor[HTML]{EFEFEF}\textbf{Full CNN} & \cellcolor[HTML]{EFEFEF}\textbf{Quantized CNN} \\ \hline
\cellcolor[HTML]{EFEFEF}\textbf{\# of Layers} & \multicolumn{2}{c|}{11 convolutional + 2 fully connected} \\ \hline
\cellcolor[HTML]{EFEFEF}\textbf{Activation Function} & \multicolumn{2}{c|}{LeakyReLU} \\ \hline
\cellcolor[HTML]{EFEFEF}\textbf{Learning Rate} & \multicolumn{2}{c|}{\(2 \times 10^{-4}\)} \\ \hline
\cellcolor[HTML]{EFEFEF}\textbf{Precision} & \texttt{float32} & \texttt{qint8} \\ \hline
\cellcolor[HTML]{EFEFEF}\textbf{Trainable Parameters} & \multicolumn{2}{c|}{$36.5 \text{M}$ }\\ \hline
\cellcolor[HTML]{EFEFEF}\textbf{Complexity (GFLOPs)} & 20.23 & 5.05 \\ \hline
\end{tabular}
\caption{CV models used for wildfire event detection}
\label{tab:cv_models}
\vspace{-10pt}
\end{table}

\subsection{Wireless Communication Network Specifications}
Table \ref{tab:network_setup} provides the configuration of the network model.

\begin{table}[H]
\centering
\resizebox{220pt}{!}{%
\begin{tabular}{|
>{\columncolor[HTML]{EFEFEF}}l |l|}
\hline
\multicolumn{1}{|c|}{\cellcolor[HTML]{EFEFEF}\textbf{Parameter}} & \multicolumn{1}{c|}{\cellcolor[HTML]{EFEFEF}\textbf{Value}} \\ \hline
\textbf{Target Signal Strength} ($tp_{k}$) & -80 dBm \\ \hline
\textbf{$\text{RC}$ Antenna Frequency} ($\mathtt{f}_{k}$) & 2.4 GHz \\ \hline
\textbf{Reference Distance} ($\mathtt{\delta_0}$) & 1 m \\ \hline
\textbf{Bandwidth min/max} ($b_{\text{min}}/b_{\text{max}}$) & 15/30 MHz \\ \hline
\textbf{Spectral Efficiency min/max }($se_{\text{min}}/se_{\text{max}}$) & 0.12/4 \\ \hline
\textbf{Noise Spectral Density} ($N_0$) & -158 dBm/Hz \\ \hline
\textbf{Path Loss Exponent} ($\mathtt{e}$) & 3.8 \\ \hline
\textbf{Shadow Fading} & 8 dB \\ \hline
\end{tabular}%
}
\caption{Network Configuration Setup}
\label{tab:network_setup}
\vspace{-8pt}
\end{table}

\subsection{Heuristic Rule-based Strategy (H-RBS)}
H-RBS performs a random-walk inspection at each timestep and determines whether to execute the task locally or offload it by comparing the transmission ratio $\frac{\mathtt{d}_{k,r}}{d_{k,r}}$ and the processing ratio $\frac{f_{k,r}}{f^{\max}_r}$, of each robot $r \in \mathcal{R}^{ready}_k$, against predefined thresholds $\text{thr}_1=0.85$ and $\text{thr}_2=0.65$. H-RBS selects the strategy whose corresponding ratio exceeds its threshold. If both exceed, the choice is random, while if neither does, the strategy with the minimum distance from its respective threshold is selected.
H-RBS integrates the early mission completion mechanism of D$^3$ARC.

\end{document}